\documentclass{article}

\usepackage{microtype}
\usepackage{graphicx}
\usepackage{subfigure}
\usepackage{booktabs} 

\usepackage{hyperref}

\usepackage[accepted]{icml2025}

\usepackage{amsmath}
\usepackage{amssymb}
\usepackage{mathtools}
\usepackage{amsthm}
\usepackage{graphics}
\usepackage{epsfig}
\usepackage{multirow}
\usepackage{xspace}
\usepackage{subfigure}
\usepackage{fancyhdr}
\usepackage{microtype}
\usepackage{color}
\usepackage{colortbl}
\usepackage{xcolor}
\definecolor{mygray}{gray}{.85}
\definecolor{myyellow}{RGB}{204,102,0}
\definecolor{myred}{RGB}{204,0,102}
\definecolor{mypurple}{RGB}{102,0,204}
\definecolor{maroon}{cmyk}{0,0.87,0.68,0.32}
\definecolor{myblue}{RGB}{227,227,240}

\usepackage{graphics}
\usepackage{amssymb}
\usepackage{amsthm}
\usepackage{mathrsfs}
\usepackage{booktabs}
\usepackage{algorithm}
\usepackage{algorithmic}
\usepackage{arydshln}
\usepackage{siunitx}
\usepackage{appendix}
\usepackage{enumitem}
\usepackage{wrapfig}

\usepackage[capitalize,noabbrev]{cleveref}

\theoremstyle{plain}

\theoremstyle{definition}

\theoremstyle{remark}

\newcommand{\model}{PPDiff\xspace}
\newcommand{\dataset}{PPBench\xspace}
\newcommand{\network}{SSINC\xspace}

\usepackage[textsize=tiny]{todonotes}

\icmltitlerunning{\model: Diffusing in Hybrid Sequence-Structure Space for Protein-Protein Complex Design}
\begin{document}

\twocolumn[

\icmltitle{\model: Diffusing in Hybrid Sequence-Structure Space for Protein-Protein Complex Design}


\icmlsetsymbol{equal}{*}

\begin{icmlauthorlist}
\icmlauthor{Zhenqiao Song}{equal,yyy}
\icmlauthor{Tianxiao Li}{comp}
\icmlauthor{Lei Li}{yyy}
\icmlauthor{Martin Renqiang Min}{comp}
\end{icmlauthorlist}
\icmlaffiliation{yyy}{Language Technologies Institute, Carnegie Mellon University}
\icmlaffiliation{comp}{NEC Laboratories America}

\icmlcorrespondingauthor{Zhenqiao Song}{zhenqiaosong@cmu.edu}

\icmlkeywords{Machine Learning, ICML}

\vskip 0.3in
]



\printAffiliationsAndNotice{\icmlEqualContribution} 

\begin{abstract}
Designing protein-binding proteins with high affinity is critical in biomedical research and biotechnology. Despite recent advancements targeting specific proteins, the ability to create high-affinity binders for arbitrary protein targets on demand, without extensive rounds of wet-lab testing, remains a significant challenge. Here, we introduce \model, a diffusion model to jointly design the sequence and structure of binders for arbitrary protein targets in a non-autoregressive manner. \model builds upon our developed \textbf{S}equence \textbf{S}tructure \textbf{I}nterleaving \textbf{N}etwork with \textbf{C}ausal attention layers~(\network), which integrates interleaved self-attention layers to capture global amino acid correlations, $k$-nearest neighbor ($k$NN) equivariant graph layers to model local interactions in three-dimensional (3D) space, and causal attention layers to simplify the intricate interdependencies within the protein sequence. To assess \model, we curate \dataset, a general protein-protein complex dataset comprising 706,360 complexes from the Protein Data Bank (PDB). The model is pretrained on \dataset and finetuned on two real-world applications: target-protein mini-binder complex design and antigen-antibody complex design. \model consistently surpasses baseline methods, achieving success rates of 50.00\%, 23.16\%, and 16.89\% for the pretraining task and the two downstream applications, respectively.
The code, data and models are available at \url{https://github.com/JocelynSong/PPDiff}.

\end{abstract}

\section{Introduction}
\label{introduction}

Designing proteins with high affinity and specificity to target proteins is important in biomedicine, with
applications spanning therapeutic development~\cite{nelson2010development},  diagnostics~\cite{brennan2010antibody} and imaging reagents~\cite{stern2013alternative}. Empirical selection methods, such as screening high-complexity random libraries of antibodies~\cite{chao2006isolating} or alternative scaffolds~\cite{hackel2008picomolar}, have demonstrated the ability to generate binders for specific targets. While effective, these approaches require considerable experimental effort, making them both time-consuming and resource-intensive. Traditional computational methods leveraging physicochemical properties offer a more efficient alternative, enabling the design of binders by focusing on specific surface patches of the protein target~\cite{chevalier2017massively,silva2019novo}.  However, these approaches are often limited by the lack of obvious surface pockets on many target proteins, thereby restricting their applicability to a narrow set of binding interactions.

In parallel with advances in traditional computational binder design, deep learning methods have achieved remarkable accuracy in predicting and designing protein-protein interactions~\cite{evans2021protein,bennett2023improving,zambaldi2024novo}. These approaches have made it possible to design binders for specific targets without relying on high-throughput screening~\cite{goudy2023silico}. In certain cases, such as peptides~\cite{vazquez2024novo} and disordered targets~\cite{wu2024sequence}, high binding affinity has been achieved without the need for extensive experimental optimization. Despite these successes, the overall success rate remains low due to two primary failure modes: the designed sequence may not fold into the intended structure, and the designed structure may not bind to the target protein. Moreover, most existing approaches are highly specialized for particular interaction types, leaving many target proteins intractable.

To address these challenges, we present \model, a diffusion model building upon our developed \textbf{S}equence \textbf{S}tructure \textbf{I}nterleaving \textbf{N}etwork with \textbf{C}ausal attention layers~(\network), to co-design the sequence and structure of binders for arbitrary protein targets in a non-autoregressive manner. In this framework, proteins are represented as residue point sets in 3D space, with each residue linked to a 3D Cartesian coordinate. The model employs a diffusion process for both continuous residue coordinates and discrete residue types, progressively adding noise to train a joint generative model. \network integrates interleaved self-attention layers, capturing global amino acid correlations, with $k$NN equivariant graph convolutional layers to model local 3D interactions. Causal attention layers are also introduced to simplify the intricate dependencies within the protein sequence. By simultaneously modeling sequence, structure, and their complex interdependencies, \model mitigates primary failure modes in binder design while improving the diversity and novelty introduced by the diffusion process to enhance success rates.

Our contributions are listed as follows:
\begin{itemize}[nosep,leftmargin=2.0em]
    \item We propose \model, a diffusion model building upon the \network network to co-design the sequence and structure of binder proteins for arbitrary protein targets. 
    \item We create \dataset, a general protein-protein complex dataset comprising 706,360 complexes curated from PDB. 
    \item Our model pretrained on \dataset demonstrates a success rate of 50.00\% on top-1 candidate, evaluated using a combination of metrics: ipTM, pTM, PAE, and pLDDT assessed by AlphaFold3~\cite{abramson2024accurate}. Notably, our model also achieves superior novelty and diversity scores, highlighting its ability to design novel and diverse protein-binding proteins with high success rates. The top-1 candidates are uploaded in the supplementary material.
    \item We finetune the pretrained \model on two important real-world applications: target-protein mini-binder complex design and antigen-antibody complex design. The finetuned models achieve success rates of 23.16\% and 16.89\%, respectively, across these tasks, demonstrating its effectiveness in addressing significant challenges in biomedicine.
    
\end{itemize}

\section{Related Work}
\label{related_work}
\textbf{Methods for Protein-Protein Complex Design.}
Designing proteins with high affinity and specificity for protein targets has been extensively explored using a wide range of approaches. The most commonly employed methods involve immunizing an animal with 
a target to induce antibody~\cite{gray2020animal} or screening high-complexity random libraries of protein scaffolds~\cite{chao2006isolating,hackel2008picomolar}. While these approaches have proven effective, they are resource-intensive and 
require significant experimental effort.  Traditional computational methods for binder design have sought to model physicochemical properties to accelerate the design process. \citet{chevalier2017massively} introduce a massively parallel approach for designing, manufacturing, and screening mini-protein binders, with biophysical property characterization.  \citet{fleishman2011computational} develop a general computational framework for designing proteins that bind to specific surface patches of targets by accounting for their physicochemical constraints. \citet{cao2022design} propose a general Rosetta-based approach to design binder proteins using only the structure of the target. While these approaches provide a systematic route for binder design, the reliance on a limited number of hotspot residues restricts them to a narrow range of interaction types. Recent advances in deep learning have significantly improved the prediction and design of protein-protein interactions~\cite{evans2021protein,humphreys2021computed,bryant2022improved,pacesa2024bindcraft}. These advancements have enabled the design of binders for certain targets without the need for high-throughput screening~\cite{watson2023novo,gainza2023novo,goudy2023silico}.  \citet{bennett2023improving} explore the augmentation of energy-based protein binder design with deep learning, while \citet{zambaldi2024novo} introduce AlphaProteo for designing binders targeting eight different proteins. Despite these developments, the overall success rate for binder design remains low, with many protein targets still intractable~\cite{yang2024design,berger2024preclinical}.

\begin{figure*}
  \centering
  \includegraphics[width=15.5cm]{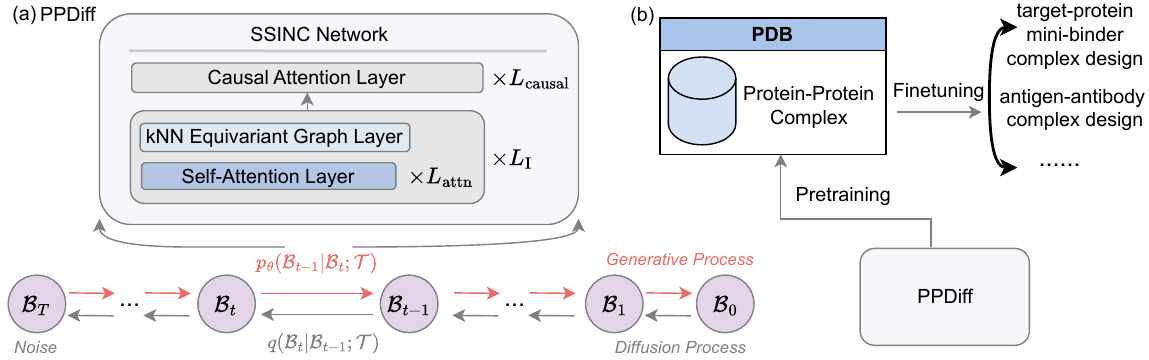}
  \vspace{-1.0em}
  \caption{
  (a) Overall architecture of our proposed \model. (b) We first pretrain \model on \dataset, a general protein-protein complex dataset curated from PDB. Then we can finetune the pretrained model on important real-world protein-protein complex design applications, such as target-protein mini-binder complex design and antigen-antibody complex design.}
  \label{Fig: model}
\end{figure*}

\textbf{Diffusion Models.} Diffusion models~\cite{sohl2015deep} have emerged as a prominent class of latent generative models. \citet{ho2020denoising} introduce denoising diffusion probabilistic models (DDPM), establishing a connection between diffusion models and denoising score-based models~\cite{song2019generative}. Diffusion models have demonstrated remarkable success in generating high-quality images~\cite{ho2020denoising,nichol2021improved} and texts~\cite{hoogeboom2021argmax,austin2021structured,li2022diffusion}. More recently, they have been applied to protein design. For instance, RFdiffusion~\cite{watson2023novo} and FrameDiff~\cite{yim2023se} are continuous diffusion models capable of generating novel protein structures. EvoDiff~\cite{alamdari2023protein} and DPLM~\cite{wangdiffusion} are discrete diffusion models designed for protein sequence generation. However, the application of diffusion models for the joint design of protein sequences and structures binding to specific protein targets still remains under-explored.

\section{Proposed Method: \model}
\label{method}

\subsection{Problem Definition}
A protein consists of a chain of amino acids connected by peptide bonds, which folds into a proper 3D structure. 
Let $\mathcal{A}$ be the set of 20 common amino acids.
Given a target protein $\mathcal{T}$ with its sequence of $M$ amino acids $\boldsymbol{s}^{\mathcal{T}}=\{s^{\mathcal{T}}_1, s^{\mathcal{T}}_2, ..., s^{\mathcal{T}}_M\} \in \mathcal{A}^M$ and the 3D coordinates of the amino acids' alpha carbons $\boldsymbol{x}^{\mathcal{T}}=[\boldsymbol{x}^{\mathcal{T}}_1, \boldsymbol{x}^{\mathcal{T}}_2, ..., \boldsymbol{x}^{\mathcal{T}}_M]^T \in R^{M\times 3}$, the goal is to generate a protein $\mathcal{B}$ that binds to the target. 
That is to predict the amino acid sequence $\boldsymbol{s}^{\mathcal{B}}=\{s^{\mathcal{B}}_1, s^{\mathcal{B}}_2, ..., s^{\mathcal{B}}_N\}\in \mathcal{A}^N$ and their alpha carbon (C$_{\alpha}$) coordinates $\boldsymbol{x}^{\mathcal{B}}=[\boldsymbol{x}^{\mathcal{B}}_1, \boldsymbol{x}^{\mathcal{B}}_2, ..., \boldsymbol{x}^{\mathcal{B}}_N]^T \in R^{N\times 3}$. 
$M$ and $N$ are the sequence lengths of the target and binder proteins. 
We formulate the protein-protein complex design problem as learning the probabilistic generative model $p_{\theta}(\mathcal{B}|\mathcal{T})$ with parameters $\theta$.
To this end, we propose a novel model \model. 


\subsection{Overall Model Architecture}
\model aims to simultaneously generate the sequence and backbone structure of a binder protein for a given protein target. 
\model is a denoising diffusion probabilistic model (DDPM)~\cite{ho2020denoising}
with a joint sequence-structure generation network (Figure~\ref{Fig: model}(a)). 
The model comprises a forward \textit{diffusion} process to add noises and a reverse \textit{generative} process, both defined as Markov chains. The diffusion process gradually perturbs the ground-truth data $\mathcal{B}_0\sim q(\mathcal{B}_0|\mathcal{T})$ into a stationary distribution $\mathcal{B}_T\sim q_{\text{noise}}$ with $T$ increasingly noisy steps:
\begin{equation}
\small
\label{forward_process}
q(\mathcal{B}_{1:T}|\mathcal{B}_0,\mathcal{T})=\Pi_{t=1}^T q(\mathcal{B}_t|\mathcal{B}_{t-1},\mathcal{T})
\end{equation}
where $\mathcal{B}_1$, ..., $\mathcal{B}_T$ is a sequence of latent variables, each contains both amino acid sequence $\boldsymbol{s}^{\mathcal{B}}_t$ and alpha carbon coordinates $\boldsymbol{x}^{\mathcal{B}}_t$.
Notice that it is a big challenge to model the latent variables because $\mathcal{B}_t$ contains both discrete and continuous components.
The generative process parameterizes the transition kernel $p_{\theta}(\mathcal{B}_{t-1}|\mathcal{B}_t, \mathcal{T})$ to recover the data distribution from the noise distribution:
\begin{equation}
\small 
\label{generative_process}
p_{\theta}(\mathcal{B}_{0:T}|\mathcal{T})=p(\mathcal{B}_T)\Pi_{t=1}^T p_{\theta}(\mathcal{B}_{t-1}|\mathcal{B}_t, \mathcal{T})
\end{equation}
To fit the model $p_{\theta}(\mathcal{B}_0|\mathcal{T})$ to the data distribution $q(\mathcal{B}_0|\mathcal{T})$, the denoising model is typically optimized by the variational bound of the log-likelihood~\cite{ho2020denoising}:
\begin{equation}
\small 
\label{diffusion_loss}
\begin{split}
&E_{q(\mathcal{B}_0|\mathcal{T})}[\log p_{\theta}(\mathcal{B}_0| \mathcal{T})]\ge E_{q(\mathcal{B}_{0:T}| \mathcal{T})}[\log \frac{p_{\theta}(\mathcal{B}_{0:T}| \mathcal{T})}{q(\mathcal{B}_{1:T}|\mathcal{B}_0, \mathcal{T})}] \\
&=E_{q(\mathcal{B}_0|\mathcal{T})}\big[\log p_{\theta}(\mathcal{B}_0|\mathcal{B}_1,\mathcal{T}) + \text{const} \\
&+\sum_{t=2}^T \underbrace{-\text{KL}[q(\mathcal{B}_{t-1}|\mathcal{B}_t, \mathcal{B}_0, \mathcal{T}) ||p_{\theta}(\mathcal{B}_{t-1}|\mathcal{B}_t, \mathcal{T})]}_{\mathcal{L}_t} \big]
\end{split}
\end{equation}
To reduce the failure modes of previous methods and enhance the consistency between the designed binder sequence and structure, \model is designed to jointly generate both the discrete residue sequence and the continuous backbone structure. This requires implementing diffusion and generative processes that operate on both discrete residue types and continuous spatial coordinates. In the following subsections, we will provide a detailed introduction of the diffusion and generative processes in \model, followed by a discussion of the learning objective and the informative guidance strategy used to generate high-quality candidates during inference.

\subsection{\model Diffusion Process}
Following \citet{guan3d}, we model the joint diffusion distribution of protein residue sequences $\boldsymbol{s}_t^{\mathcal{B}}$ and backbone structures 
$\boldsymbol{x}^{\mathcal{B}}_t$ independently. 
Such a formulation allows for efficient sampling of noisy data for both modalities, i.e. discrete residue types and continuous coordinates. For the discrete protein sequence $\boldsymbol{s}^{\mathcal{B}}$, we adopt a categorical distribution $\text{Cat}(\boldsymbol{s}^{\mathcal{B}};\boldsymbol{p})$ as proposed by \citet{hoogeboom2021argmax}, where $\boldsymbol{p}$ represents a vector on the ($|\mathcal{A}|-1=19$)-dimensional probability simplex. For the continuous backbone structure, we follow \citet{ho2020denoising} and use a Gaussian distribution $\mathcal{N}$ to model spatial coordinates. At each diffusion step $t$, uniform noise is added to the residue types across all categories, and Gaussian noise is applied to the structural coordinates. This process follows a Markov chain with a predefined schedule $\beta_1$, $\beta_2$, ..., $\beta_T$ as follows:
\begin{equation}
\small 
\label{diffusion_process_time_step}
\begin{split}
q(\mathcal{B}_t|\mathcal{B}_{t-1}, \mathcal{T}) &= q(\boldsymbol{s}^{\mathcal{B}}_t|\boldsymbol{s}^{\mathcal{B}}_{t-1}, \mathcal{T})\cdot q(\boldsymbol{x}^{\mathcal{B}}_t|\boldsymbol{x}^{\mathcal{B}}_{t-1}, \mathcal{T}) \\
q(\boldsymbol{s}^{\mathcal{B}}_t|\boldsymbol{s}^{\mathcal{B}}_{t-1}, \mathcal{T}) &= \text{Cat}(\boldsymbol{s}^{\mathcal{B}}_t;(1-\beta_t)\boldsymbol{s}^{\mathcal{B}}_{t-1}+\beta_t/K) \\
q(\boldsymbol{x}^{\mathcal{B}}_t|\boldsymbol{x}^{\mathcal{B}}_{t-1}, \mathcal{T}) &=\mathcal{N}(\boldsymbol{x}^{\mathcal{B}}_t;\sqrt{1-\beta_t}\boldsymbol{x}^{\mathcal{B}}_{t-1},\beta_t\boldsymbol{I})
\end{split}
\end{equation}
In practice, the schedules for different data modalities can be different. 

Denoting $\alpha_t=1-\beta_t$ and $\overline{\alpha}_t=\Pi_{i=1}^t\alpha_i$, we can calculate the noisy data distributions $q(\boldsymbol{s}^{\mathcal{B}}_t|\boldsymbol{s}^{\mathcal{B}}_{0}, \mathcal{T})$ and $q(\boldsymbol{x}^{\mathcal{B}}_t|\boldsymbol{x}^{\mathcal{B}}_{0}, \mathcal{T}) $ in closed-form:
\begin{equation}
\small 
\label{diffusion_process_time0}
\begin{split}
q(\boldsymbol{s}^{\mathcal{B}}_t|\boldsymbol{s}^{\mathcal{B}}_{0}, \mathcal{T}) &= \text{Cat}(\boldsymbol{s}^{\mathcal{B}}_t;\overline{\alpha}_t\boldsymbol{s}^{\mathcal{B}}_{0}+(1-\overline{\alpha}_t)/K)\\
q(\boldsymbol{x}^{\mathcal{B}}_t|\boldsymbol{x}^{\mathcal{B}}_{0}, \mathcal{T}) &=\mathcal{N}(\boldsymbol{x}^{\mathcal{B}}_t;\sqrt{\overline{\alpha}_t}\boldsymbol{x}^{\mathcal{B}}_{0}, (1-\overline{\alpha}_t)\boldsymbol{I})
\end{split}
\end{equation}
Based on Eq.~\ref{diffusion_process_time_step} and Eq.~\ref{diffusion_process_time0}, we can calculate the posterior distributions $q(\boldsymbol{s}^{\mathcal{B}}_{t-1}|\boldsymbol{s}^{\mathcal{B}}_{t}, \boldsymbol{s}^{\mathcal{B}}_{0}, \mathcal{T})$ and $q(\boldsymbol{x}^{\mathcal{B}}_{t-1}|\boldsymbol{x}^{\mathcal{B}}_{t}, \boldsymbol{x}^{\mathcal{B}}_{0}, \mathcal{T})$ using Bayes rule in closed-form:
\begin{equation}
\small 
\label{posterior_distribution}
\begin{split}
&q(\boldsymbol{s}^{\mathcal{B}}_{t-1}|\boldsymbol{s}^{\mathcal{B}}_{t}, \boldsymbol{s}^{\mathcal{B}}_{0}, \mathcal{T}) = \text{Cat}(\boldsymbol{s}^{\mathcal{B}}_{t-1};\boldsymbol{\theta}_{\text{post}}^{\boldsymbol{s}}(\boldsymbol{s}^{\mathcal{B}}_{t}, \boldsymbol{s}^{\mathcal{B}}_{0}))\\
&\boldsymbol{\theta}_{\text{post}}^{\boldsymbol{s}}(\boldsymbol{s}^{\mathcal{B}}_{t}, \boldsymbol{s}^{\mathcal{B}}_{0}) = \Tilde{\boldsymbol{\theta}}^{\boldsymbol{s}} / \sum_{k=1}^{|\mathcal{A}|} \Tilde{\theta}_k^{\boldsymbol{s}} \\
&\Tilde{\boldsymbol{\theta}}^{\boldsymbol{s}} = [\alpha_t\boldsymbol{s}_t^{\mathcal{B}}+(1-\alpha_t)/K]\odot [\overline{\alpha}_{t-1}\boldsymbol{s}^{\mathcal{B}}_0+(1-\overline{\alpha}_{t-1})/K] \\
&q(\boldsymbol{x}^{\mathcal{B}}_{t-1}|\boldsymbol{x}^{\mathcal{B}}_{t}, \boldsymbol{x}^{\mathcal{B}}_{0}, \mathcal{T})=\mathcal{N}(\boldsymbol{x}^{\mathcal{B}}_{t-1};\Tilde{\boldsymbol{\mu}}_t(\boldsymbol{x}^{\mathcal{B}}_t, \boldsymbol{x}^{\mathcal{B}}_0), \Tilde{\beta}_t\boldsymbol{I})\\
&\Tilde{\boldsymbol{\mu}}_t(\boldsymbol{x}^{\mathcal{B}}_t, \boldsymbol{x}^{\mathcal{B}}_0)=\frac{\sqrt{\overline{\alpha}_{t-1}}\beta_t}{1-\overline{\alpha}_t}\boldsymbol{x}^{\mathcal{B}}_0+\frac{\sqrt{\alpha_t}(1-\overline{\alpha}_{t-1})}{1-\overline{\alpha}_t}\boldsymbol{x}^{\mathcal{B}}_t \\
&\Tilde{\beta}_t=\frac{1-\overline{\alpha}_{t-1}}{1-\overline{\alpha}_t} \beta_t
\end{split}
\end{equation}

\subsection{The \network Network for \model Generative Process}
The generative process recovers the data distribution from the noise distribution. We parameterize the reverse generative process using our developed \network network by $\theta$:
\begin{equation}
\small 
\label{generative_process}
\begin{split}
&p_{\theta}(\mathcal{B}_{t-1}|\mathcal{B}_t, \mathcal{T})=p_{\theta}(\boldsymbol{s}^{\mathcal{B}}_{t-1}|\boldsymbol{s}^{\mathcal{B}}_{t}, \boldsymbol{x}^{\mathcal{B}}_{t}, \mathcal{T}) \cdot p_{\theta}(\boldsymbol{x}^{\mathcal{B}}_{t-1}|\boldsymbol{x}^{\mathcal{B}}_{t}, \boldsymbol{s}^{\mathcal{B}}_{t}, \mathcal{T})  \\
&p_{\theta}(\boldsymbol{s}^{\mathcal{B}}_{t-1}|\boldsymbol{s}^{\mathcal{B}}_{t}, \boldsymbol{x}^{\mathcal{B}}_{t}, \mathcal{T}) = \text{Cat}(\boldsymbol{s}^{\mathcal{B}}_{t-1};\boldsymbol{\theta}_{\text{post}}^{\boldsymbol{s}}(\boldsymbol{s}^{\mathcal{B}}_{t}, \boldsymbol{\hat{s}}^{\mathcal{B}}_{0})) \\
&p_{\theta}(\boldsymbol{x}^{\mathcal{B}}_{t-1}|\boldsymbol{x}^{\mathcal{B}}_{t}, \boldsymbol{s}^{\mathcal{B}}_{t}, \mathcal{T}) = \mathcal{N}(\boldsymbol{x}^{\mathcal{B}}_{t-1};\boldsymbol{\mu}_{\theta}( \boldsymbol{s}^{\mathcal{B}}_t, \boldsymbol{x}^{\mathcal{B}}_t, t, \mathcal{T}), \sigma^2_t\boldsymbol{I})
\end{split}
\end{equation}
The \network network is composed of $L_{\text{I}}$ interleaved blocks, each containing $L_{\text{attn}}$ self-attention layers~\cite{vaswani2017attention} to capture global correlations among amino acids and a $k$NN equivariant graph convolutional layer~\cite{satorras2021n} to model local interactions among neighboring residues in 3D space, as illustrated in Figure~\ref{Fig: model} (a). In the $l$-th interleaved block, the representations are calculated as follows:
\begin{equation}
\small
\begin{split}
&\boldsymbol{H}_t^{l+0.5}=\text{Self-Attention}(\boldsymbol{H}_t^{l};\mathcal{T}) \\
&\boldsymbol{m}_{t,ik}^{l+0.5} = f_m([\boldsymbol{h}_{t,i}^{l+0.5}; \boldsymbol{h}_{t,k}^{l+0.5}; ||\boldsymbol{x}_{t,i}^{l}-\boldsymbol{x}^{l}_{t,k}||_2];\mathcal{T})\\
&w_{t,ik}^{l+0.5} = \mathrm{Softmax}(\boldsymbol{m}_{t,ik}^{l+0.5}),\boldsymbol{m}_{t,ik}^{l+1} = w_{t,ik}^{l+0.5} \cdot \boldsymbol{m}_{t,ik}^{l+0.5}\\
&\boldsymbol{c}_{i}^{l+1} = \sum\nolimits_{k\in \mathrm{N(i)}} \boldsymbol{m}_{t,ik}^{l+1}, \mathrm{weight}=\sigma(f_{w}(\boldsymbol{c}_{i}^{l+1})) \\
&\boldsymbol{h}_{t,i}^{l+1} = \boldsymbol{h}_{t,i}^{l+0.5} +  \sum\nolimits_{k\in \mathrm{N(i)}} \mathrm{weight}\cdot\boldsymbol{m}_{t,ik}^{l+1}\\ 
&\boldsymbol{x}^{l+1}_{t,i}=\boldsymbol{x}^{l}_{t,i}+\sum\nolimits_{k\in \mathrm{N(i)}}(\boldsymbol{x}^{l}_{t,i}-\boldsymbol{x}^{l}_{t,k})\cdot f_x(\boldsymbol{m}_{t,ik}^{l+1})\\
\end{split}
\end{equation}
where $\boldsymbol{H}_t^l=[\boldsymbol{h}_{t,1}^l,\boldsymbol{h}_{t,2}^l, ..., \boldsymbol{h}_{t,N}^l]^T$ denotes the residue representation matrix at the $l$-th block and time step $t$, while $\boldsymbol{H}_t^0=[\boldsymbol{s}^{\mathcal{B}}_1,\boldsymbol{s}^{\mathcal{B}}_2,...,\boldsymbol{s}^{\mathcal{B}}_N]^T$ is the embedding matrix of binder protein sequence. Functions $f_*$ denote feed-forward layers and $\sigma$ denotes sigmoid function. $N(i)$ is the set of $k$-nearest neighbors of the $i$-th residue. Finally, $\boldsymbol{x}^{L_{\text{I}}}_t$ is the predicted structure at time step $t$, i.e. $\boldsymbol{\hat{x}}^{\mathcal{B}}_0$.
To make the learning easier, $L_{\text{causal}}$ causal attention layers are added on top of the \network network to capture the sequential dependencies:
\begin{equation}
\small
\begin{split}
&\boldsymbol{h}_{t,i}^{\text{out}}=\text{Causal-Attention} \left(\boldsymbol{h}_{t,i}^{L_{\text{I}}}, \boldsymbol{H}_{t,1:i}^{L_{\text{I}}} \right)\\
&\boldsymbol{\hat{s}}^{\mathcal{B}}_0=\mathrm{Softmax}(\boldsymbol{H}^{\mathrm{out}}_t)
\end{split}
\label{Eq: casual_attention_layer}
\end{equation}
where the $i$-th residue can only attend to its previous residues $\boldsymbol{H}_{t,1:i}^{L_{\text{I}}}$. 
We can then approximate the posterior distributions $p_{\theta}(\boldsymbol{s}^{\mathcal{B}}_{t-1}|\boldsymbol{s}^{\mathcal{B}}_{t}, \boldsymbol{x}^{\mathcal{B}}_{t}, \mathcal{T})$ and $p_{\theta}(\boldsymbol{x}^{\mathcal{B}}_{t-1}|\boldsymbol{x}^{\mathcal{B}}_{t}, \boldsymbol{s}^{\mathcal{B}}_{t}, \mathcal{T})$ using the predicted $\boldsymbol{\hat{s}}^{\mathcal{B}}_0$ and $\boldsymbol{\hat{x}}^{\mathcal{B}}_0$.

\subsection{Training Objective}
We learn the model by maximizing the variational lower bound of the log-likelihood defined in Eq.~\ref{diffusion_loss}, where $\mathcal{L}_t$ is calculated as:
\begin{equation}
\small
\label{training_objective}
\begin{split}
\mathcal{L}_t&=-\text{KL}[q(\boldsymbol{s}^{\mathcal{B}}_{t-1}|\boldsymbol{s}^{\mathcal{B}}_{t},\boldsymbol{s}^{\mathcal{B}}_{0},\mathcal{T})||p_{\theta}(\boldsymbol{s}^{\mathcal{B}}_{t-1}|\boldsymbol{s}^{\mathcal{B}}_{t},\boldsymbol{x}^{\mathcal{B}}_{t}, \mathcal{T})]\\&-\text{KL}[q(\boldsymbol{x}^{\mathcal{B}}_{t-1}|\boldsymbol{x}^{\mathcal{B}}_{t},\boldsymbol{x}^{\mathcal{B}}_{0},\mathcal{T})||p_{\theta}(\boldsymbol{x}^{\mathcal{B}}_{t-1}|\boldsymbol{x}^{\mathcal{B}}_{t},\boldsymbol{s}^{\mathcal{B}}_{t}, \mathcal{T})] \\
&=\underbrace{-\left\{\sum_k \boldsymbol{\theta}_{\text{post}}^{\boldsymbol{s}}(\boldsymbol{s}^{\mathcal{B}}_{t}, \boldsymbol{s}^{\mathcal{B}}_{0})_k \log \frac{\boldsymbol{\theta}_{\text{post}}^{\boldsymbol{s}}(\boldsymbol{s}^{\mathcal{B}}_{t}, \boldsymbol{s}^{\mathcal{B}}_{0})_k}{\boldsymbol{\theta}_{\text{post}}^{\boldsymbol{s}}(\boldsymbol{s}^{\mathcal{B}}_{t}, \boldsymbol{\hat{s}}^{\mathcal{B}}_{0})_k}\right\}}_{\mathcal{L}_t^{\boldsymbol{s}}} \\
&\underbrace{-\left\{\frac{1}{2\sigma^2_t}||\Tilde{\boldsymbol{\mu}}_t(\boldsymbol{x}^{\mathcal{B}}_t, \boldsymbol{x}^{\mathcal{B}}_0)-\boldsymbol{\mu}_{\theta}( \boldsymbol{s}^{\mathcal{B}}_t, \boldsymbol{x}^{\mathcal{B}}_t, t, \mathcal{T})||^2+\text{const}\right\}}_{\mathcal{L}_t^{\boldsymbol{x}}}
\end{split}
\end{equation}
where $\boldsymbol{\theta}_{\text{post}}^{\boldsymbol{s}}(\boldsymbol{s}^{\mathcal{B}}_{t}, \boldsymbol{\hat{s}}^{\mathcal{B}}_{0})$ is defined in Eq.~\ref{posterior_distribution}, with $\boldsymbol{\hat{s}}^{\mathcal{B}}_{0}$ calculated in Eq.~\ref{Eq: casual_attention_layer}. $\mathcal{L}_t^{\boldsymbol{x}}$ can be further symplified as:
\begin{equation}
\small
\mathcal{L}_t^{\boldsymbol{x}}=-\left\{\lambda_t||\boldsymbol{x}^{\mathcal{B}}_0-\boldsymbol{\hat{x}}^{\mathcal{B}}_0||^2 + \text{const}\right\}, \quad \lambda_t=\frac{\overline{\alpha}_{t-1}\beta_t^2}{2\sigma^2(1-\overline{\alpha}_{t})^2}
\end{equation}
Following previous work~\cite{guan3d}, we set $\lambda_t=1$. Furthermore,  $\log p_{\theta}(\mathcal{B}_0|\mathcal{B}_1,\mathcal{T})$ can be calculated as :
\begin{equation}
\small 
\begin{split}
\log p_{\theta}(\mathcal{B}_0|\mathcal{B}_1,\mathcal{T})&=\log p_{\theta}(\boldsymbol{s}^{\mathcal{B}}_0|\boldsymbol{s}^{\mathcal{B}}_1, \boldsymbol{x}^{\mathcal{B}}_{1}, \mathcal{T}) \\&+ \log p_{\theta}(\boldsymbol{x}^{\mathcal{B}}_0|\boldsymbol{x}^{\mathcal{B}}_1, \boldsymbol{s}^{\mathcal{B}}_{1}, \mathcal{T}) \\
&=\sum_k \boldsymbol{s}^{\mathcal{B}}_{0,k}\log\boldsymbol{\hat{s}}^{\mathcal{B}}_{0,k} - ||\boldsymbol{x}^{\mathcal{B}}_0-\boldsymbol{\hat{x}}^{\mathcal{B}}_0||^2
\end{split}
\end{equation}

\subsection{Informative Guidance for Inference}
As suggested by previous studies~\cite{graikos2022diffusion,wu2022diffusion}, an informative prior distribution can significantly enhance model performance. To leverage this insight, we introduce an informative guidance to select better starting points rather than random noise. For backbone structure guidance, we define the following energy function:
\begin{equation}
\small 
E_{\text{knn}}(\mathcal{B})=\sum_{i=1}^N\left(\text{knn-dist}(\boldsymbol{x}_i)-\mu_{\mathrm{knn}}\right)^2
\end{equation}
where $\text{knn-dist}(\boldsymbol{x}_i)=\frac{1}{k}\sum_{j\in N(i)}||\boldsymbol{x}_i-\boldsymbol{x}_j||^2$ denotes the average distance from $i$-th residue to its $k$ nearest neighbors and $\mu_{\text{knn}}$ is the empirical mean of $\text{knn-dist}(\boldsymbol{x})$ across all residues in the training dataset. In our setting, $k$ is set to 4. During inference, we sample 10 noisy structures and choose the one with the lowest energy. For sequence guidance, we randomly sample secondary structure fragments from the training dataset, identified by using DSSP~\cite{kabsch1983dictionary}. This strategy ensures the initialized structures and sequences maintain geometric similarities to the training data, providing the model with a more informed starting point and enhancing the quality of candidate generation.

\section{Experiments}
\label{experiments}
In this section, we first describe our \dataset construction process in Section~\ref{Sec:dataset_construction} and experimental setup in Section~\ref{Sec:experimental_setup}. Then we conduct extensive experiments on a \textbf{General Protein-Protein Complex Design} task (Section~\ref{Sec: generation_protein_complex_design}) and two real-world applications, \textbf{Target-Protein Mini-Binder Complex Design} (Section~\ref{Sec:binder_design}) and \textbf{Antigen-Antibody Complex Design} (Section~\ref{Sec:antibody_design}). The specific experimental settings are introduced in each task section.

\subsection{\dataset Construction}
\label{Sec:dataset_construction}
Our goal is to develop a general protein-protein complex design model that serves as a foundational 
model, adaptable for finetuning in diverse real-world protein-protein interaction applications. 
To achieve this, we curate a large-scale protein-protein complex dataset by identifying chain-pair interfaces, following the data processing pipeline in AlphaFold3~(AF3). Details of the data curation process are provided in Appendix~\ref{Appendix:protein_protein_complex curation_detail}. This process results in 367,016 chain-pair interfaces. Treating each chain in a chain-pair interface as a potential target protein, we gather a total of 734,032 complexes. Protein sequences are then clustered based on 50\% sequence identity, producing 22,847 clusters. To prepare the dataset for training and evaluation, we designate 10 clusters each for validation and testing, with the remaining clusters reserved for training. To ensure efficient processing, training data are further filtered to exclude sequences longer than 1,024 residues, while validation and testing data are restricted to sequences no longer than 512 residues, finally resulting in a total of 706,360 complexes. Detailed data statistics  are provided in the Appendix~\ref{Appendix: data_statistics_procomplexbench}.   

\subsection{Experimental Setup}
\label{Sec:experimental_setup}
\textbf{Implementation Details.}
Our architecture incorporates three interleaving blocks and one causal attention layer. Each interleaving block comprises 11 self-attention layers and one $k$NN-based equivariant graph convolutional layer. The parameters for self-attention layers are initialized using the pretrained 650M ESM-2 model~\cite{lin2022language}. \model has a total of 692M parameters. The diffusion process is configured with 1,000 steps, employing a cosine schedule for sequence diffusion and a sigmoid schedule for structure diffusion. Additional model training details are provided in Appendix~\ref{Appendix: model_training_details}.

\textbf{Baseline Models.}
Given the absence of large-scale models for protein-protein complex sequence and structure co-design, we evaluate the effectiveness of \model by comparing it with several representative baselines:
(1) \textbf{SEnc+ProteinMPNN} first employs a continuous diffusion model based on structured encoder~\cite{ingraham2019generative} enhanced with spatial features to design complex backbone structures, followed by ProteinMPNN~\cite{dauparas2022robust} to predict binder protein sequences based on the designed backbone structure.
(2) \textbf{InterleavingDiff} is a variant of \model without the casual attention layers, of which the backbone model is the interleaving network~\cite{songgenerative}.
(3) \textbf{\network Network} has the same architecture as \model, but it is trained without diffusion process. To ensure a fair comparison, all models are trained on the curated \dataset using their official implementations.

\begin{table*}[!t]
\footnotesize
\begin{center}
\setlength{\tabcolsep}{1.8mm}{
\begin{tabular}{llccccccc}
\midrule
\multicolumn{2}{c}{Methods}  & ipTM ($\uparrow$) & pTM ($\uparrow$) & PAE ($\downarrow$) & pLDDT($\uparrow$) & Success Rate ($\uparrow$) & Novelty ($\uparrow$) & Diversity ($\uparrow$) \\
\midrule
\multicolumn{2}{c}{Ground Truth} &0.691& 0.782 & 7.901& 86.987&69.64\%&-- &-- \\
\hdashline
\multirow{4}{*}{top-1}&SEnc +ProteinMPNN&0.629&0.713&10.374&79.819&35.08\%&59.76\%&-- \\
& InterleavingDiff &0.674&0.757&9.934&82.411&45.61\%&89.45\% & --\\
&\network Network&0.660&0.751&9.713&82.998&47.36\%&37.46\%& --\\
&\cellcolor{myblue}\model &\cellcolor{myblue}\textbf{0.700}&\cellcolor{myblue}\textbf{0.779}&\cellcolor{myblue}\textbf{9.153}&\cellcolor{myblue}\textbf{83.765}&\cellcolor{myblue}\textbf{50.00\%}&\cellcolor{myblue}\textbf{89.46\%}&\cellcolor{myblue}-- \\
\midrule
\multirow{4}{*}{top-5}&SEnc +ProteinMPNN&0.617&	0.682&	11.789&78.021&	29.82\%&	61.78\%	&58.72\% \\
& InterleavingDiff &0.648&	0.732&	11.010&	79.944&	37.85\%	&\textbf{90.66\%}	&\textbf{91.82\%}\\
&\network Network&0.647&0.745&\textbf{9.903}&\textbf{82.290}&	43.85\%&	37.49\%&	15.12\% \\
&\cellcolor{myblue}\model &\cellcolor{myblue}\textbf{0.665}&\cellcolor{myblue}\textbf{0.747}&\cellcolor{myblue}10.231&\cellcolor{myblue}81.659&\cellcolor{myblue}\textbf{45.71\%}&\cellcolor{myblue}88.93\%&\cellcolor{myblue}90.76\% \\
\midrule
\multirow{4}{*}{top-10}&SEnc +ProteinMPNN&0.582&0.671
&12.347&76.892&21.05\%&62.98\%&	58.93\% \\
& InterleavingDiff &0.629&0.716&11.745&	78.328&24.28\%&\textbf{90.64\%}&\textbf{91.89\%}\\
&\network Network&0.620&0.721&\textbf{10.075}&\textbf{81.852}&	36.84\%&37.45\%&15.03\% \\
&\cellcolor{myblue}\model &\cellcolor{myblue}\textbf{0.633}&\cellcolor{myblue}\textbf{0.729}&\cellcolor{myblue}10.895&\cellcolor{myblue}80.322&\cellcolor{myblue}\textbf{37.68\%}&\cellcolor{myblue}89.10\%&\cellcolor{myblue}91.09\% \\
\bottomrule
\end{tabular}}
\end{center}
\vspace{-1.2em}
\caption{Model performance on general protein-protein complex design task. ($\uparrow$): the higher the better. ($\downarrow$): the lower the better. Calculating the diversity score for the top-1 candidate is unnecessary, as it consists of only a single candidate. \model consistently achieves the highest success rates for top-1, top-5, and top-10 candidates out of 100 candidates for each target protein.}
\label{Table:general_protein_complex_design}
\end{table*}

\subsection{General Protein-Protein Complex Design}
\label{Sec: generation_protein_complex_design}
This task aims to develop a general protein-protein complex design foundational model that can be readily finetuned for diverse real-world protein-protein interaction applications (Figure~\ref{Fig: model} (b)). To this end, we pretrain our proposed \model on the curated \dataset.

\textbf{Evaluation Metrics.}
We evaluate the designed protein-protein complexes based on structure stability, reliability, and functional interaction using AF3 metrics. Specifically, \textbf{pLDDT} assesses structural stability, \textbf{pTM} and \textbf{PAE} measure structural reliability, and \textbf{ipTM} evaluates interface interactions, reflecting functionality. Detailed explanations of these metrics are provided in Appendix~\ref{Appendix:metrics_explanation}. As per prior studies~\cite{binder2022alphafold,bennett2023improving,abramson2024accurate}, a successful complex satisfies ipTM$\ge$0.8, pTM$\ge$0.8, PAE$\le$10 and pLDDT$\ge$80. For each target protein $\mathcal{T}$ in the test set $\mathcal{D}_{\text{test}}$, we generate 100 binder candidates and rank them by their complex pLDDT scores, selecting the top-$k$ candidates $\mathcal{C}$. The \textbf{success rate (SR)} for the top-$k$ candidates is:
\begin{equation}
\small 
\mathrm{SR}=\frac{1}{|\mathcal{D}_{\text{test}}| * k}\sum_{\mathcal{T} \in \mathcal{D}_{\text{test}}} \sum_{i=1}^k \mathrm{II}(f(\mathcal{T}, \mathcal{B}_i))
\end{equation}
where $f(\mathcal{T}, \mathcal{B}_i)$ ensures if the thresholds for ipTM, pTM, PAE, and pLDDT are met, and $\mathrm{II(true)}=1$, $\mathrm{II(false)}=0$. Here, $\mathcal{B}_i$ is the $i$-th candidate.
We also evaluate \textbf{diversity} and \textbf{novelty} scores. Diversity quantifies sequence variability among the top-$k$ candidates, while novelty measures the deviation of candidates from the ground-truth sequence:
\begin{equation}
\small 
\begin{split}
\mathrm{diversity}&=\frac{1}{(k*(k-1))} \sum_{i,j\in \mathcal{C}, i\neq j} 1-\mathrm{AAR}(\boldsymbol{s}^{\mathcal{B}}_i, \boldsymbol{s}^{\mathcal{B}}_j)\\
\mathrm{novelty}&=\frac{1}{k}\sum_{i\in \mathcal{C}}1-\mathrm{AAR(\boldsymbol{s}^{\mathcal{B}}_i, \boldsymbol{s}^{\mathcal{B}})}
\end{split}
\end{equation}
where $\boldsymbol{s}^{\mathcal{B}}_i$ is the sequence of $\mathcal{B}_i$, $\boldsymbol{s}^{\mathcal{B}}$ is the ground truth binder sequence, and AAR denotes the amino acid recovery rate. The final diversity and novelty scores are averaged across the whole test set.

\textbf{Main Results.} 
The results for top-1, top-5, and top-10 candidates are reported in Table~\ref{Table:general_protein_complex_design}. \textbf{These results demonstrate that \model excels in designing high-quality, diverse, and novel protein-binding proteins across a wide range of protein targets.} For the top-1 candidate, \model achieves the best performance across all metrics. While \model performs slightly worse than the \network Network on PAE and pLDDT for top-5 and top-10 candidates, it still achieves the highest success rate and significantly performs better in diversity and novelty. Although the diversity and novelty scores for \model's top-5 and top-10 candidates are marginally lower than those of InterleavingDiff, \model produces designs of much higher quality, validating the effectiveness of the causal attention layer.

\begin{table*}[!t]
\small
\begin{center}
\setlength{\tabcolsep}{1.25mm}{
\begin{tabular}{lccccccccccc}
\midrule
\multirow{2}{*}{Models} & \multicolumn{5}{c}{Seen Class} & \multicolumn{5}{c}{Zero-Shot} & \multirow{2}{*}{Average} \\
\cmidrule(r){2-6} \cmidrule(r){7-11} 
& FGFR2	& InsulinR	& PDGFR	& TGFb &	VirB8&	H3	& IL7Ra&EGFR&	TrkA&	Tie2 &  \\
\midrule
SEnc+ProteinMPNN&8.07\%&	6.08\%&	4.61\%&	15.56\%&	11.42\%&	4.73\%&	17.14\%&	0.0	&5.00\%&	0.0&	7.12\%\\
InterleavingDiff & 10.00\%&	6.95\%&	11.53\%&	35.56\%&	8.57\%&	48.42\%&	34.28\%&	0.0	&10.00\%&	10.00\%&	19.32\%\\
\network Network &\textbf{21.05\%} &	0.0&	3.07\%&	20.00\%&	2.85\%	&7.36\%&	22.85\%&	0.0&	5.00\%&	0.0&	10.96\%\\
\rowcolor{myblue}
\model &7.36\%&	\textbf{10.43\%}&	\textbf{14.61\%}&	\textbf{35.56\%}&	\textbf{11.42\%}&	\textbf{55.26\%}&	\textbf{60.00\%}	&0.0&	\textbf{20.00\%}&	\textbf{30.00\%}&	\textbf{23.16\%}\\
\bottomrule
\end{tabular}}
\end{center}
\vspace{-1.2em}
\caption{The success rate (\textbf{\%}, $\uparrow$) for all models on the target-protein mini-binder complex design task. ``Average" refers to the overall success rate across the entire test set, rather than a simple mean across target categories. Our proposed \model demonstrates a significant improvement in success rate, outperforming all previous methods by a substantial margin.}
\label{Table:target_protein_minibinder_design_success_rate}
\end{table*}

\begin{table*}[!t]
\footnotesize
\begin{center}
\setlength{\tabcolsep}{0.88mm}{
\begin{tabular}{lcccccccc}
\midrule
Methods  & ipTM ($\uparrow$) & pTM ($\uparrow$) & PAE ($\downarrow$) & pLDDT($\uparrow$) & Success Rate ($\uparrow$) & H1 Novelty ($\uparrow$) & H2 Novelty ($\uparrow$) & H3 Novelty ($\uparrow$)\\
\midrule
SEnc +ProteinMPNN&0.448	&0.578	&14.418	&77.786&	6.24\%&	\textbf{61.78\%}&	59.53\%&	70.85\%\\
InterleavingDiff &0.496&	0.653&	13.583	&80.939	&12.00\%&	57.54\%&	63.59\%&	75.58\%\\
\network Network&0.539&	0.654&	\textbf{12.064}	&82.706	&16.74\%&	30.44\%&	37.70\%&	59.86\%\\
\rowcolor{myblue}
\model & \textbf{0.541}&	\textbf{0.668}&	12.999	&\textbf{82.827}&	\textbf{16.89\%}&	57.79\%&	\textbf{66.39\%}&	\textbf{76.17\%}\\
\bottomrule
\end{tabular}}
\end{center}
\vspace{-1.2em}
\caption{Model performance on antigen-antibody complex design task. ($\uparrow$): the higher the better. ($\downarrow$): the lower the better. Novelty scores for CDR-H1, CDR-H2, and CDR-H3 are denoted as H1, H2, and H3 novelty, respectively. \model demonstrates superior performance, achieving the highest average success rate across the designed CDRs.}
\label{Table:antibody_antigen_complex_design}
\end{table*}

\subsection{Target-Protein Mini-Binder Complex Design}
\label{Sec:binder_design}
In this section, we aim to use \model to design proteins that bind to specific protein targets with high affinity.

\textbf{Datasets.} 
We collect experimentally confirmed positive target-protein mini-binder complexes against ten targets with diverse structural properties from \citet{bennett2023improving}. For categories containing more than 50 complexes, an 8:1:1 random split is applied for training, validation, and test sets. For smaller categories, all complexes are included in the test set to create a zero-shot evaluation scenario. Detailed data statistics are provided in Appendix~\ref{Appendix: data_statistics_for_target_protein_mini_binder}.

\textbf{Evaluation Metrics.} 
Following prior works~\cite{bennett2023improving,watson2023novo,songsurfpro}, we evaluate the binding affinities between the designed binders and target proteins using the \textbf{AlphaFold2 (AF2) pAE\_interaction}\footnote{https://github.com/nrbennet/dl\_binder\_design} developed by \citet{bennett2023improving}. Their work demonstrates that AF2 pAE\_interaction effectively differentiates experimentally validated binders from non-binders, achieving success rates ranging from 1.5\% and 7\% for target proteins FGFR2, IL7Ra, TrkA, InsulinR, PDGFR, and VirB8. Based on the calculated pAE\_interaction score, we determine the \textbf{success rate} by generating five mini-binder candidates for each target protein. The success rate is defined as the proportion of designed binders that achieve a lower pAE\_interaction score with the target protein than the ground truth positive binder. More evaluation details are provided in Appendix~\ref{Appendix: evaluation-detail-of-target-protein-minibinder-complex}.

\textbf{Main Results.} We finetune \model on the target-protein mini-binder complex design task and the success rates are summarized in Table~\ref{Table:target_protein_minibinder_design_success_rate}. \textbf{Our \model achieves a superior success rate compared to all baselines, demonstrating its effectiveness in designing high-affinity binders.} However, we observe that all models fail to design successful binders for EGFR, likely due to the lack of deep pockets or distinct features in its binding sites~\cite{callaway2024ai}. 

\subsection{Antigen-Antibody Complex Design}
\label{Sec:antibody_design}
In this section, we apply \model to design high-quality antibodies binding to given antigens.

\textbf{Datasets.}
Following \citet{jiniterative}, we curate antigen-antibody complexes from the Structural Antibody Database~(\citet{dunbar2014sabdab}, SAbDab), resulting in 4,261 valid complexes after removing cases lacking a light chain or antigen. The dataset is split into training, validation, and test sets based on the clustering of CDRs. Taking CDR-H3 as an example, we use MMseqs2~\cite{steinegger2017mmseqs2} to cluster CDR-H3 sequences at 40\% sequence identity, calculated using the BLOSUM62 substitution matrix~\cite{henikoff1992amino}. Clusters are then randomly divided into training, validation, and test sets in an 8:1:1 ratio. The same procedure is applied to create splits for CDR-H1 and CDR-H2, resulting in 1,034, 1,418, and 2,254 clusters, respectively. Details of the training, validation and test set sizes for each CDR clustering are provided in the Appendix~\ref{Appendix:antibody-antigen-complex-data-statistics}.

\textbf{Evaluation Metrics.}
To guarantee a comprehensive evaluation of the designed antigen-antibody complex, we use the \textbf{ipTM, pTM, PAE, pLDDT, success rate and novelty} metrics introduced in the general protein-protein complex design evaluation~(Section~\ref{Sec: generation_protein_complex_design}).

\textbf{Main Results.}
We finetune the pretrained \model on each of the antigen-antibody complex dataset derived from the clustering process and present the average performance in Table~\ref{Table:antibody_antigen_complex_design}. Detailed results for each CDR clustering are given in Appendix~\ref{Appendix:addtional-result-antibody-antigen}. \textbf{\model achieves the highest antigen-antibody binding affinity (ipTM), success rate, and novelty scores.} It demonstrates \model's ability to design high-affinity and novel antibodies for given antigens, highlighting its potential significance for real-world applications.

\section{Analysis: Diving Deep into \model}
\label{analysis}
\begin{figure*}
\begin{minipage}[t]{0.25\linewidth}
\centering
\includegraphics[width=4.0cm]{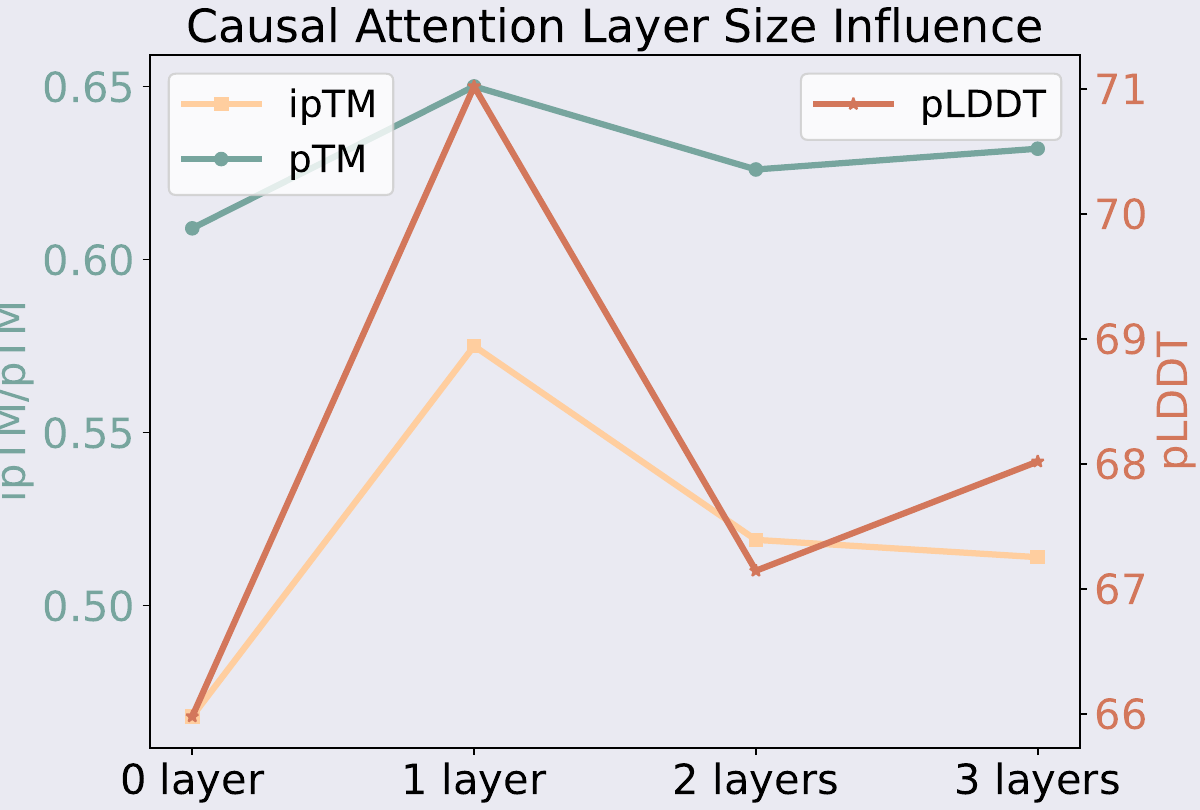}
\centerline{\small{(a) Causal Attention Layer Size}}
\end{minipage}%
\begin{minipage}[t]{0.25\linewidth}
\centering
\includegraphics[width=4.05cm]{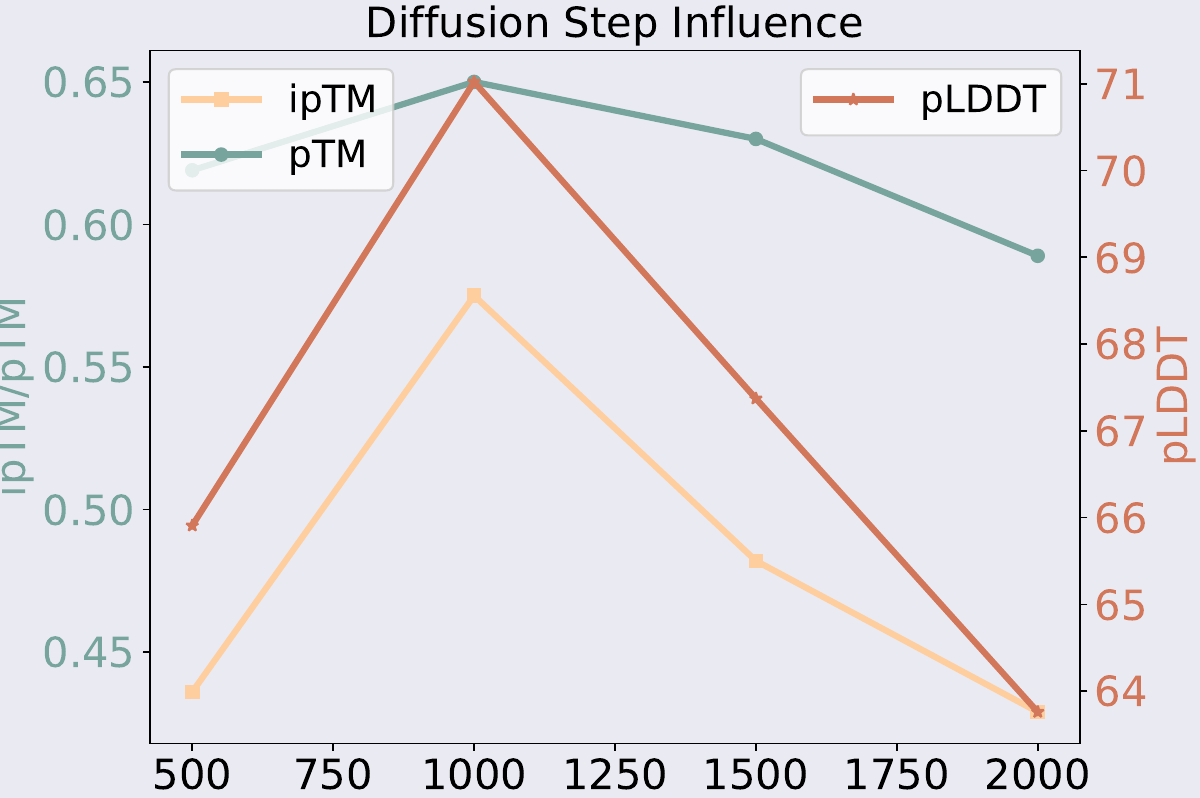}
\centerline{\small{(b) Diffusion Step}}
\end{minipage}%
\begin{minipage}[t]{0.25\linewidth}
\centering
\includegraphics[width=3.55cm]{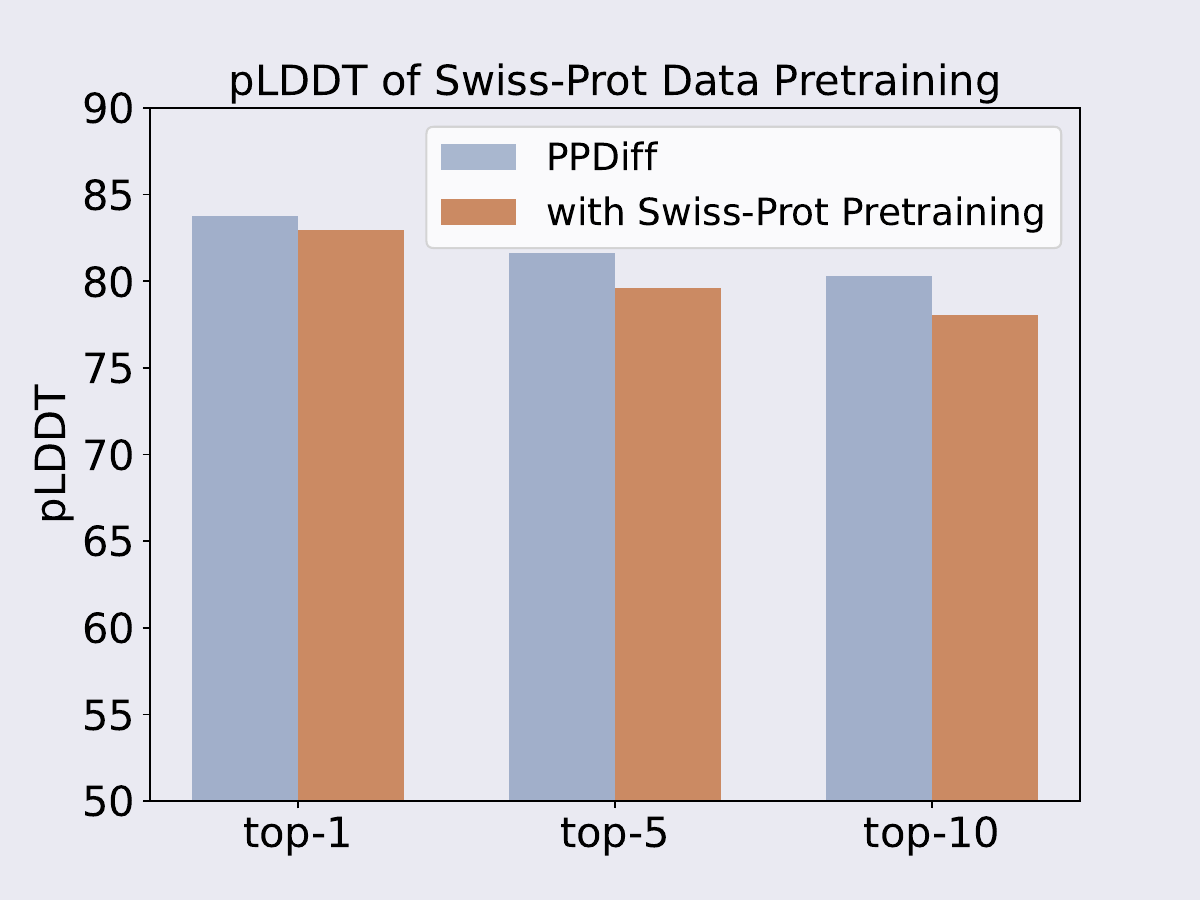}
\centerline{\small{(c) pLDDT of Pretraining}}
\end{minipage}%
\begin{minipage}[t]{0.25\linewidth}
\centering
\includegraphics[width=3.55cm]{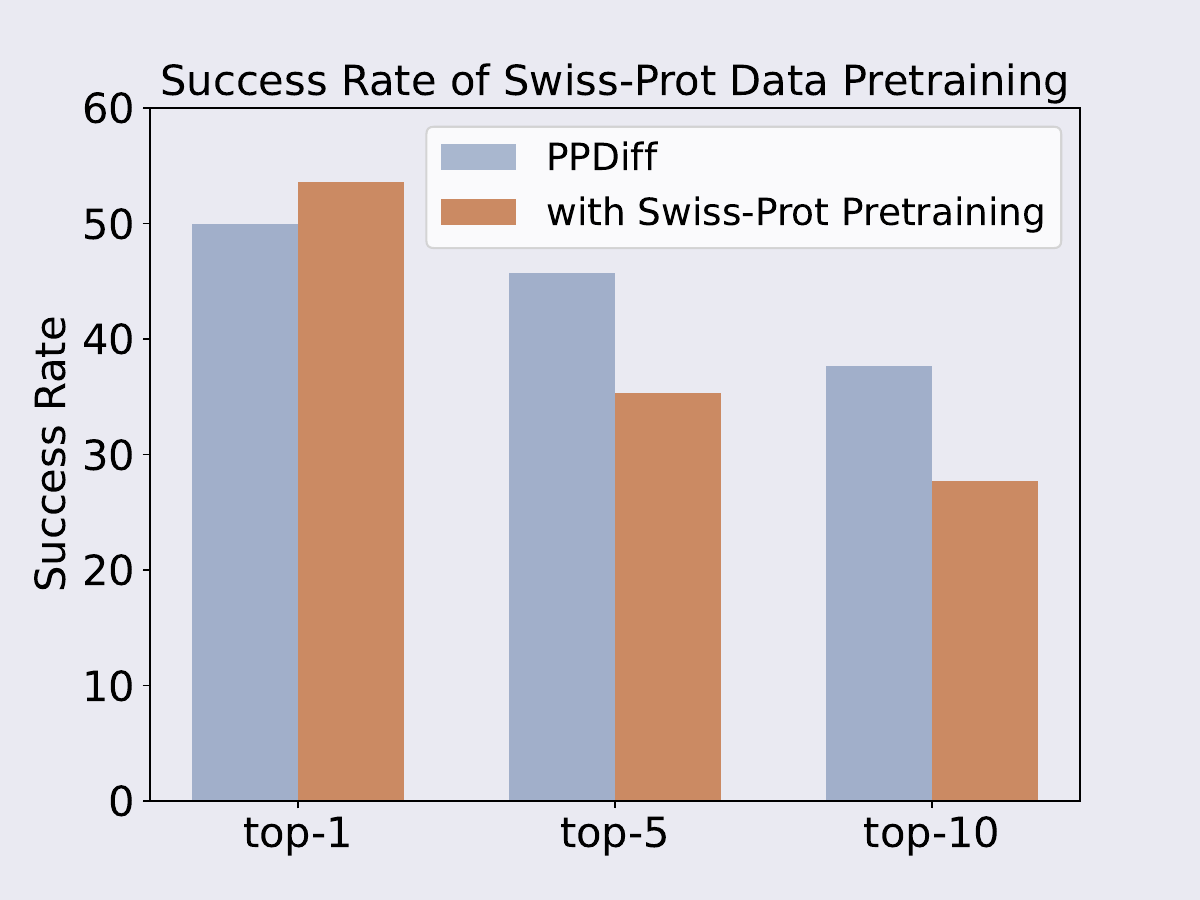}
\centerline{\small{(d) Success Rate of Pretraining}}
\end{minipage}
\begin{minipage}[t]{0.25\linewidth}
\centering
\includegraphics[width=4.05cm]{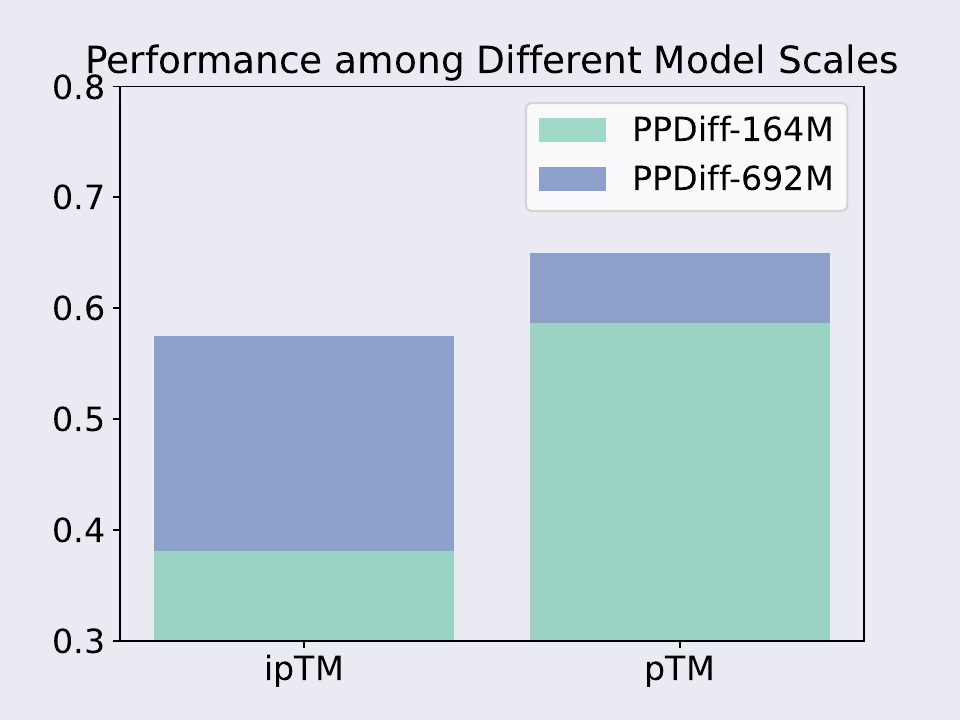}
\centerline{\small{(e) Model Scaling}}
\end{minipage}%
\begin{minipage}[t]{0.25\linewidth}
\centering
\includegraphics[width=3.95cm]{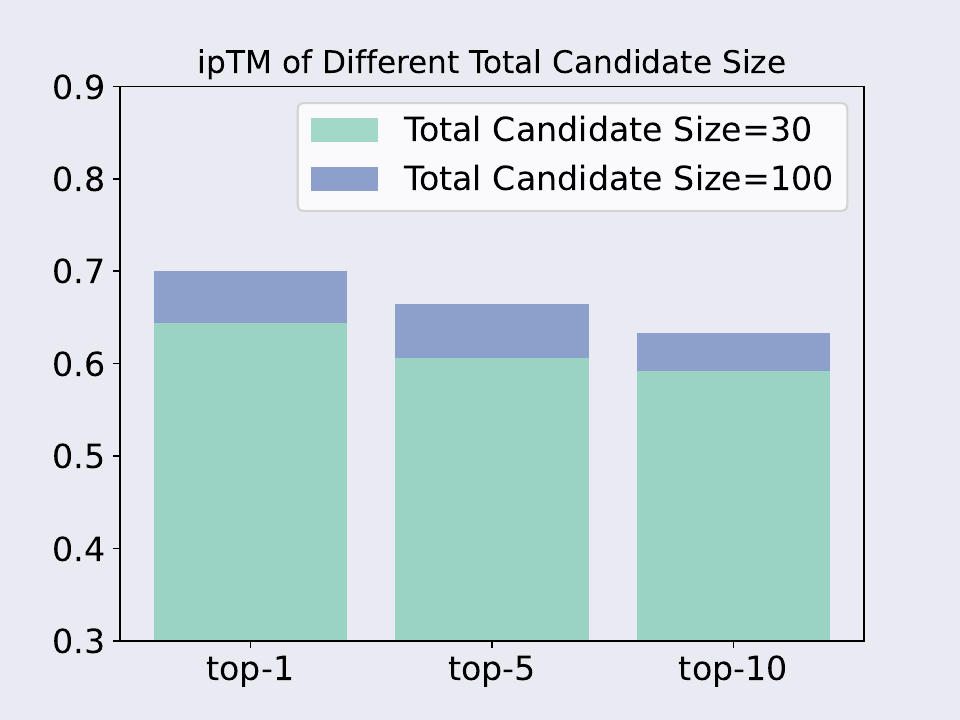}
\centerline{\small{(f) ipTM score}}
\end{minipage}%
\begin{minipage}[t]{0.25\linewidth}
\centering
\includegraphics[width=3.9cm]{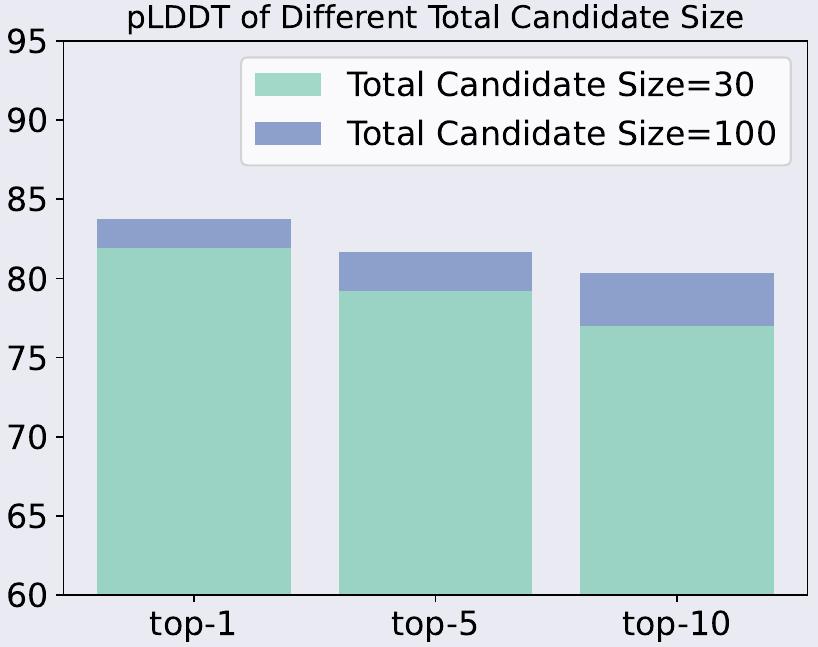}
\centerline{\small{(g) pLDDT score}}
\end{minipage}%
\begin{minipage}[t]{0.25\linewidth}
\centering
\includegraphics[width=3.9cm]{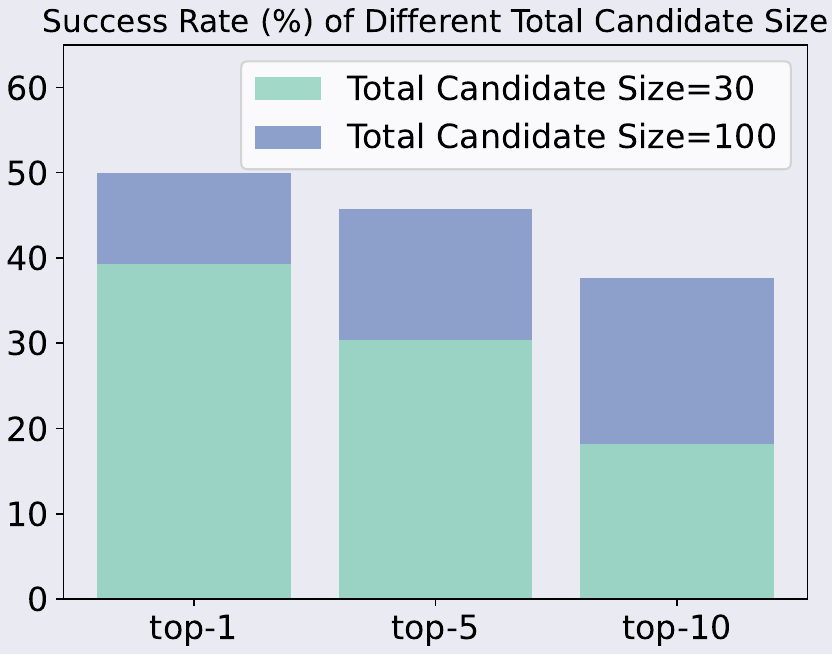}
\centerline{\small{(h) Success Rate (\%)}}
\end{minipage}
\vspace{-1.0em}
	\caption{Ablation study on: (a) causal attention layer size, (b) different diffusion steps, (c-d) additional Swiss-Prot data pretraining, (e) different model scales, (f-h) different total candidate sizes.} 
 \label{Fig: ablation_study}
\end{figure*}

\begin{figure*}
  \centering
  \includegraphics[width=17.0cm]{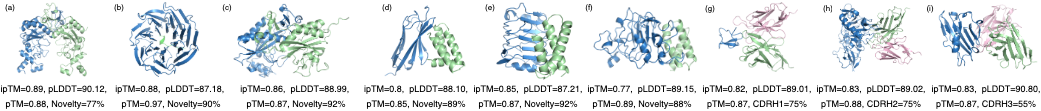}
  \vspace{-0.8em}
  \caption{Designed protein complexes from: (a-c) general protein-protein complex design, (d-f) target-protein mini-binder complex design, and (g-i) antigen-antibody complex design. Target proteins are shown in blue, and the designed binder proteins in green, with light chains in pink for antigen-antibody complexes. \model is able to design high-affinity protein-binding proteins across diverse target scaffolds.}
  \label{Fig: case_study}
\end{figure*}

\subsection{How Does The Causal Attention Layer Function?}
To assess the influence of causal attention layers, we train the model with varying layer size (0 to 4) on \dataset. For each setting, we generate one candidate and evaluate the resulting complexes using AF3 with the same seed. The results, visualized in Figure~\ref{Fig: ablation_study} (a), illustrate that introducing causal attention layers significantly enhances design quality. However, increasing the number of layers above one does not provide additional benefits. Thus, \textbf{using a single causal attention layer achieves the optimal design quality}.

\subsection{Effect of Diffusion Steps on Model Performance}
To explore the impact of diffusion steps on \model's performance, we train the model with different diffusion steps ranging from 500 to 2,000. For each trained model, a single candidate is generated and evaluated using AF3 under the same seed. The results, presented in Figure~\ref{Fig: ablation_study} (b), demonstrate that \textbf{\model performs best when configured with 1,000 diffusion steps}.

\subsection{Does Model Scale Help?}
\textbf{\model is scalable.} To evaluate the scalability of \model, we compare the performance of our \model (692M) with a version comprising 30 self-attention layers (164M). For each trained model, a single candidate is generated and evaluated using AF3 under the same seed. Figure~\ref{Fig: ablation_study} (e) shows the ipTM and pTM scores. The results reveal that the scaled-up \model achieves significantly higher ipTM and pTM scores, highlighting the scalability and enhanced performance of our approach as the model size increases.

\subsection{Impact of Additional Swiss-Prot Data Pretraining}
To assess whether pretraining on additional data improves design quality, we pretrain \model on the Swiss-Prot dataset~\cite{boeckmann2003swiss} and its AlphaFold-derived structures~\cite{varadi2022alphafold}. Since Swiss-Prot proteins are monomers, pseudo complexes are generated by identifying residue pairs within 5Å in the 3D structure and with sequence length longer than 60, designating the interacting residues as the last and first residues of two chains. The model is first pretrained on this pseudo complex dataset before continuing training on \dataset. Figure~\ref{Fig: ablation_study} (c) and (d) compare the pLDDT and success rates of top-1, top-5, and top-10 candidates from this model with those of the original \model. \textbf{The results indicate that pretraining on additional Swiss-Prot data provides no significant benefits.} Detailed results are included in the Appendix~\ref{Appendix: Siwss-Prot-pretraining}.

\subsection{Influence of Total Candidate Size}
To evaluate the impact of total candidate size, we generate 30 binder proteins for each target protein in the test set of \dataset and analyze the top-1, top-5, and top-10 results, as shown in Figure~\ref{Fig: ablation_study} (f-h). Detailed results are provided in Appendix~\ref{Appendix: candidate-pool-size}. The results indicate that increasing the total candidate size to 100 significantly enhances the quality of top-$k$ candidates ($k$=1,5, and 10), particularly in terms of success rate. This demonstrates that \model can produce higher-quality complexes when considering a larger pool of candidates, which is potentially advantageous for wet-lab validation in real-world applications.

\begin{table*}[!t]
\small
\begin{center}
\setlength{\tabcolsep}{1.25mm}{
\begin{tabular}{lccccccccccc}
\midrule
\multirow{2}{*}{Models} & \multicolumn{5}{c}{Seen Class} & \multicolumn{5}{c}{Zero-Shot} & \multirow{2}{*}{Average} \\
\cmidrule(r){2-6} \cmidrule(r){7-11} 
& FGFR2	& InsulinR	& PDGFR	& TGFb &	VirB8&	H3	& IL7Ra&EGFR&	TrkA&	Tie2 &  \\
\midrule
SEnc+ProteinMPNN&-197.82	&-192.56&	-231.46	&-203.41&	-235.67&	-198.23	&-192.85&	-178.23	&-224.91&	-201.32	&-205.64\\
InterleavingDiff&-230.40&	-233.49&	-234.60&	-231.93	&-222.56&	-227.03&	-229.93	&-218.26	&-234.12&	-230.43&	-230.25\\
\network Network &-207.76&	-193.18	&-226.77&	-211.04	&-220.34&	-206.02&	-207.24&	-183.53	&-217.91&	-196.22	&-208.48\\
\rowcolor{myblue}
\model &\textbf{-256.86}&	\textbf{-260.95	}&	\textbf{-270.55	}&	\textbf{-251.35}&	\textbf{-252.69	}&	\textbf{-244.23}&	\textbf{-261.36}	&\textbf{-244.75}&	\textbf{-266.06}&	\textbf{-265.19}&	\textbf{-256.45}\\
\bottomrule
\end{tabular}}
\end{center}
\vspace{-1.2em}
\caption{The docking score ($\downarrow$) for all models on the target-protein mini-binder complex design task. ``Average" refers to the overall docking score across the entire test set, rather than a simple mean across target categories. Our proposed \model demonstrates the best binding affinity across all targets, outperforming all baseline methods.}
\label{Table:target_protein_minibinder_design_docking}
\end{table*}

\begin{table*}[!t]
\small
\begin{center}
\setlength{\tabcolsep}{0.7mm}{
\begin{tabular}{lccccccccccccc}
\midrule
\multirow{2}{*}{Models} & \multicolumn{5}{c}{Seen Class} & \multicolumn{5}{c}{Zero-Shot} & \multirow{2}{*}{Average} &\multirow{2}{*}{Nov.} &\multirow{2}{*}{Div.} \\
\cmidrule(r){2-6} \cmidrule(r){7-11} 
& FGFR2	& InsulinR	& PDGFR	& TGFb &	VirB8&	H3	& IL7Ra&EGFR&	TrkA&	Tie2 &  \\
\midrule
RF+MPNN &\textbf{28.07\%}&	8.69\%&	\textbf{15.38\%}&	22.22\%&	\textbf{57.14\%}&	7.89\%&	28.57\%&	\textbf{25.00\%}&	\textbf{100.00\%}&	0.0&	21.46\%&78.10\%	&25.71\%\\
\rowcolor{myblue}
\model &7.36\%&	\textbf{10.43\%}&	14.61\%&	\textbf{35.56\%}&	11.42\%&	\textbf{55.26\%}&	\textbf{60.00\%}	&0.0&	20.00\%&	\textbf{30.00\%}&	\textbf{23.16\%}&\textbf{91.39\%}&	\textbf{91.79\%}\\
\bottomrule
\end{tabular}}
\end{center}
\vspace{-1.2em}
\caption{The success rate (\textbf{\%}, $\uparrow$) for \model and a strong baseline RFDiffusion+ProteinMPNN (called RF+MPNN here due to the limited space) on the target-protein mini-binder complex design task. ``Average" refers to the overall success rate across the entire test set, rather than a simple mean across target categories. Our \model achieves a higher average success rate, as well as significantly higher novelty~(Nov.) and diversity~(Div.) scores.}
\label{Table:target_protein_minibinder_design_success_rate_rfdiffusion}
\end{table*}

\subsection{In-Depth Exploration of Target Protein–Mini Binder Complex Design}
To better evaluate the quality of our designed binders, we first compute the docking scores between the target proteins and the designed binders using HDOCK~\cite{yan2020hdock}. The results, summarized in Table~\ref{Table:target_protein_minibinder_design_docking}, show that \model achieves better docking performance to existing baseline methods, exhibiting similar phenomenon using AF2\_pAE interaction scores. This validates both the reliability of the AF2 pAE\_interaction score as a proxy for binder quality and the ability of \model to generate binders with strong binding affinity to the target proteins.

We further compare \model against a strong baseline, RFDiffusion combined with ProteinMPNN (RF+MPNN), with results presented in Table~\ref{Table:target_protein_minibinder_design_success_rate_rfdiffusion}. Our model achieves a higher average success rate across the 10 target proteins. Moreover, the binders generated by \model exhibit significantly greater diversity and novelty compared to those produced by the baseline. These findings highlight the effectiveness of \model in designing functional, diverse, and novel protein binders.

\subsection{Designing Novel and Diverse Protein Complexes}
Figure~\ref{Fig: case_study} showcases nine designed protein-protein complexes across different design tasks: (a-c) general protein-protein complex design, (d-f) target-protein mini-binder complex design, and (g-i) antigen-antibody complex design. The target proteins in these examples exhibit diverse structural scaffolds, and all complexes achieve outstanding metrics, including ipTM scores approaching 0.8 or higher, pTM scores exceeding 0.8, pLDDT values above 80, and novelty scores surpassing 50\%. These results highlight \textbf{\model's capability to design novel, high-affinity binder proteins for a wide range of protein targets}. More designed complexes are provided in Appendix~\ref{Appendix: case_study}.

\section{Conclusion}
\label{conclusion}
In this paper, we present \model, a diffusion model building upon \network network to co-design binder protein sequences and structures for specified protein targets. \network incorporates interleaved self-attention layers and $k$NN-based equivariant graph layers, complemented by causal attention layers, enabling it to jointly model sequences, structures, and their complicated interdependencies. The model is pretrained on a general protein-protein complex design task and finetuned for two critical real-world applications, achieving success rates of 50.00\%, 23.16\% and 16.89\%, respectively. A potential limitation of this work is the absence of wet-lab validation, which is an essential step planned for our future research to further validate the model's practical utility.

\section*{Impact Statement}
This paper presents work whose goal is to advance the field of 
Machine Learning. There are many potential societal consequences 
of our work, none of which we feel must be specifically highlighted here.

\bibliography{icml2025}

\begin{thebibliography}{57}
\providecommand{\natexlab}[1]{#1}
\providecommand{\url}[1]{\texttt{#1}}
\expandafter\ifx\csname urlstyle\endcsname\relax
  \providecommand{\doi}[1]{doi: #1}\else
  \providecommand{\doi}{doi: \begingroup \urlstyle{rm}\Url}\fi

\bibitem[Abramson et~al.(2024)Abramson, Adler, Dunger, Evans, Green, Pritzel, Ronneberger, Willmore, Ballard, Bambrick, et~al.]{abramson2024accurate}
Abramson, J., Adler, J., Dunger, J., Evans, R., Green, T., Pritzel, A., Ronneberger, O., Willmore, L., Ballard, A.~J., Bambrick, J., et~al.
\newblock Accurate structure prediction of biomolecular interactions with alphafold 3.
\newblock \emph{Nature}, pp.\  1--3, 2024.

\bibitem[Alamdari et~al.(2023)Alamdari, Thakkar, van~den Berg, Tenenholtz, Strome, Moses, Lu, Fusi, Amini, and Yang]{alamdari2023protein}
Alamdari, S., Thakkar, N., van~den Berg, R., Tenenholtz, N., Strome, B., Moses, A., Lu, A.~X., Fusi, N., Amini, A.~P., and Yang, K.~K.
\newblock Protein generation with evolutionary diffusion: sequence is all you need.
\newblock \emph{BioRxiv}, pp.\  2023--09, 2023.

\bibitem[Austin et~al.(2021)Austin, Johnson, Ho, Tarlow, and Van Den~Berg]{austin2021structured}
Austin, J., Johnson, D.~D., Ho, J., Tarlow, D., and Van Den~Berg, R.
\newblock Structured denoising diffusion models in discrete state-spaces.
\newblock \emph{Advances in neural information processing systems}, 34:\penalty0 17981--17993, 2021.

\bibitem[Bennett et~al.(2023)Bennett, Coventry, Goreshnik, Huang, Allen, Vafeados, Peng, Dauparas, Baek, Stewart, et~al.]{bennett2023improving}
Bennett, N.~R., Coventry, B., Goreshnik, I., Huang, B., Allen, A., Vafeados, D., Peng, Y.~P., Dauparas, J., Baek, M., Stewart, L., et~al.
\newblock Improving de novo protein binder design with deep learning.
\newblock \emph{Nature Communications}, 14\penalty0 (1):\penalty0 2625, 2023.

\bibitem[Berger et~al.(2024)Berger, Seeger, Yu, Aydin, Yang, Rosenblum, Guenin-Mac{\'e}, Glassman, Arguinchona, Sniezek, et~al.]{berger2024preclinical}
Berger, S., Seeger, F., Yu, T.-Y., Aydin, M., Yang, H., Rosenblum, D., Guenin-Mac{\'e}, L., Glassman, C., Arguinchona, L., Sniezek, C., et~al.
\newblock Preclinical proof of principle for orally delivered th17 antagonist miniproteins.
\newblock \emph{Cell}, 187\penalty0 (16):\penalty0 4305--4317, 2024.

\bibitem[Binder et~al.(2022)Binder, Berendzen, Stevens, He, Wang, Dokholyan, and Oprea]{binder2022alphafold}
Binder, J.~L., Berendzen, J., Stevens, A.~O., He, Y., Wang, J., Dokholyan, N.~V., and Oprea, T.~I.
\newblock Alphafold illuminates half of the dark human proteins.
\newblock \emph{Current Opinion in Structural Biology}, 74:\penalty0 102372, 2022.

\bibitem[Boeckmann et~al.(2003)Boeckmann, Bairoch, Apweiler, Blatter, Estreicher, Gasteiger, Martin, Michoud, O'Donovan, Phan, et~al.]{boeckmann2003swiss}
Boeckmann, B., Bairoch, A., Apweiler, R., Blatter, M.-C., Estreicher, A., Gasteiger, E., Martin, M.~J., Michoud, K., O'Donovan, C., Phan, I., et~al.
\newblock The swiss-prot protein knowledgebase and its supplement trembl in 2003.
\newblock \emph{Nucleic acids research}, 31\penalty0 (1):\penalty0 365--370, 2003.

\bibitem[Brennan et~al.(2010)Brennan, O'connor, Rexhepaj, Ponten, and Gallagher]{brennan2010antibody}
Brennan, D.~J., O'connor, D.~P., Rexhepaj, E., Ponten, F., and Gallagher, W.~M.
\newblock Antibody-based proteomics: fast-tracking molecular diagnostics in oncology.
\newblock \emph{Nature Reviews Cancer}, 10\penalty0 (9):\penalty0 605--617, 2010.

\bibitem[Bryant et~al.(2022)Bryant, Pozzati, and Elofsson]{bryant2022improved}
Bryant, P., Pozzati, G., and Elofsson, A.
\newblock Improved prediction of protein-protein interactions using alphafold2.
\newblock \emph{Nature communications}, 13\penalty0 (1):\penalty0 1265, 2022.

\bibitem[Callaway(2024)]{callaway2024ai}
Callaway, E.
\newblock Ai has dreamt up a blizzard of new proteins. do any of them actually work?
\newblock \emph{Nature}, 634\penalty0 (8034):\penalty0 532--533, 2024.

\bibitem[Cao et~al.(2022)Cao, Coventry, Goreshnik, Huang, Sheffler, Park, Jude, Markovi{\'c}, Kadam, Verschueren, et~al.]{cao2022design}
Cao, L., Coventry, B., Goreshnik, I., Huang, B., Sheffler, W., Park, J.~S., Jude, K.~M., Markovi{\'c}, I., Kadam, R.~U., Verschueren, K.~H., et~al.
\newblock Design of protein-binding proteins from the target structure alone.
\newblock \emph{Nature}, 605\penalty0 (7910):\penalty0 551--560, 2022.

\bibitem[Chao et~al.(2006)Chao, Lau, Hackel, Sazinsky, Lippow, and Wittrup]{chao2006isolating}
Chao, G., Lau, W.~L., Hackel, B.~J., Sazinsky, S.~L., Lippow, S.~M., and Wittrup, K.~D.
\newblock Isolating and engineering human antibodies using yeast surface display.
\newblock \emph{Nature protocols}, 1\penalty0 (2):\penalty0 755--768, 2006.

\bibitem[Chevalier et~al.(2017)Chevalier, Silva, Rocklin, Hicks, Vergara, Murapa, Bernard, Zhang, Lam, Yao, et~al.]{chevalier2017massively}
Chevalier, A., Silva, D.-A., Rocklin, G.~J., Hicks, D.~R., Vergara, R., Murapa, P., Bernard, S.~M., Zhang, L., Lam, K.-H., Yao, G., et~al.
\newblock Massively parallel de novo protein design for targeted therapeutics.
\newblock \emph{Nature}, 550\penalty0 (7674):\penalty0 74--79, 2017.

\bibitem[Dauparas et~al.(2022)Dauparas, Anishchenko, Bennett, Bai, Ragotte, Milles, Wicky, Courbet, de~Haas, Bethel, et~al.]{dauparas2022robust}
Dauparas, J., Anishchenko, I., Bennett, N., Bai, H., Ragotte, R.~J., Milles, L.~F., Wicky, B.~I., Courbet, A., de~Haas, R.~J., Bethel, N., et~al.
\newblock Robust deep learning--based protein sequence design using proteinmpnn.
\newblock \emph{Science}, 378\penalty0 (6615):\penalty0 49--56, 2022.

\bibitem[Dunbar et~al.(2014)Dunbar, Krawczyk, Leem, Baker, Fuchs, Georges, Shi, and Deane]{dunbar2014sabdab}
Dunbar, J., Krawczyk, K., Leem, J., Baker, T., Fuchs, A., Georges, G., Shi, J., and Deane, C.~M.
\newblock Sabdab: the structural antibody database.
\newblock \emph{Nucleic acids research}, 42\penalty0 (D1):\penalty0 D1140--D1146, 2014.

\bibitem[Evans et~al.(2021)Evans, O’Neill, Pritzel, Antropova, Senior, Green, {\v{Z}}{\'\i}dek, Bates, Blackwell, Yim, et~al.]{evans2021protein}
Evans, R., O’Neill, M., Pritzel, A., Antropova, N., Senior, A., Green, T., {\v{Z}}{\'\i}dek, A., Bates, R., Blackwell, S., Yim, J., et~al.
\newblock Protein complex prediction with alphafold-multimer.
\newblock \emph{biorxiv}, pp.\  2021--10, 2021.

\bibitem[Fleishman et~al.(2011)Fleishman, Whitehead, Ekiert, Dreyfus, Corn, Strauch, Wilson, and Baker]{fleishman2011computational}
Fleishman, S.~J., Whitehead, T.~A., Ekiert, D.~C., Dreyfus, C., Corn, J.~E., Strauch, E.-M., Wilson, I.~A., and Baker, D.
\newblock Computational design of proteins targeting the conserved stem region of influenza hemagglutinin.
\newblock \emph{Science}, 332\penalty0 (6031):\penalty0 816--821, 2011.

\bibitem[Gainza et~al.(2023)Gainza, Wehrle, Van Hall-Beauvais, Marchand, Scheck, Harteveld, Buckley, Ni, Tan, Sverrisson, et~al.]{gainza2023novo}
Gainza, P., Wehrle, S., Van Hall-Beauvais, A., Marchand, A., Scheck, A., Harteveld, Z., Buckley, S., Ni, D., Tan, S., Sverrisson, F., et~al.
\newblock De novo design of protein interactions with learned surface fingerprints.
\newblock \emph{Nature}, 617\penalty0 (7959):\penalty0 176--184, 2023.

\bibitem[Goudy et~al.(2023)Goudy, Nallathambi, Kinjo, Randolph, and Kuhlman]{goudy2023silico}
Goudy, O.~J., Nallathambi, A., Kinjo, T., Randolph, N.~Z., and Kuhlman, B.
\newblock In silico evolution of autoinhibitory domains for a pd-l1 antagonist using deep learning models.
\newblock \emph{Proceedings of the National Academy of Sciences}, 120\penalty0 (49):\penalty0 e2307371120, 2023.

\bibitem[Graikos et~al.(2022)Graikos, Malkin, Jojic, and Samaras]{graikos2022diffusion}
Graikos, A., Malkin, N., Jojic, N., and Samaras, D.
\newblock Diffusion models as plug-and-play priors.
\newblock \emph{Advances in Neural Information Processing Systems}, 35:\penalty0 14715--14728, 2022.

\bibitem[Gray et~al.(2020)Gray, Bradbury, Knappik, Pl{\"u}ckthun, Borrebaeck, and D{\"u}bel]{gray2020animal}
Gray, A., Bradbury, A.~R., Knappik, A., Pl{\"u}ckthun, A., Borrebaeck, C.~A., and D{\"u}bel, S.
\newblock Animal-free alternatives and the antibody iceberg.
\newblock \emph{Nature Biotechnology}, 38\penalty0 (11):\penalty0 1234--1239, 2020.

\bibitem[Guan et~al.(2023)Guan, Qian, Peng, Su, Peng, and Ma]{guan3d}
Guan, J., Qian, W.~W., Peng, X., Su, Y., Peng, J., and Ma, J.
\newblock 3d equivariant diffusion for target-aware molecule generation and affinity prediction.
\newblock In \emph{The Eleventh International Conference on Learning Representations}, 2023.

\bibitem[Hackel et~al.(2008)Hackel, Kapila, and Wittrup]{hackel2008picomolar}
Hackel, B.~J., Kapila, A., and Wittrup, K.~D.
\newblock Picomolar affinity fibronectin domains engineered utilizing loop length diversity, recursive mutagenesis, and loop shuffling.
\newblock \emph{Journal of molecular biology}, 381\penalty0 (5):\penalty0 1238--1252, 2008.

\bibitem[Henikoff \& Henikoff(1992)Henikoff and Henikoff]{henikoff1992amino}
Henikoff, S. and Henikoff, J.~G.
\newblock Amino acid substitution matrices from protein blocks.
\newblock \emph{Proceedings of the National Academy of Sciences}, 89\penalty0 (22):\penalty0 10915--10919, 1992.

\bibitem[Ho et~al.(2020)Ho, Jain, and Abbeel]{ho2020denoising}
Ho, J., Jain, A., and Abbeel, P.
\newblock Denoising diffusion probabilistic models.
\newblock \emph{Advances in neural information processing systems}, 33:\penalty0 6840--6851, 2020.

\bibitem[Hoogeboom et~al.(2021)Hoogeboom, Nielsen, Jaini, Forr{\'e}, and Welling]{hoogeboom2021argmax}
Hoogeboom, E., Nielsen, D., Jaini, P., Forr{\'e}, P., and Welling, M.
\newblock Argmax flows and multinomial diffusion: Learning categorical distributions.
\newblock \emph{Advances in Neural Information Processing Systems}, 34:\penalty0 12454--12465, 2021.

\bibitem[Humphreys et~al.(2021)Humphreys, Pei, Baek, Krishnakumar, Anishchenko, Ovchinnikov, Zhang, Ness, Banjade, Bagde, et~al.]{humphreys2021computed}
Humphreys, I.~R., Pei, J., Baek, M., Krishnakumar, A., Anishchenko, I., Ovchinnikov, S., Zhang, J., Ness, T.~J., Banjade, S., Bagde, S.~R., et~al.
\newblock Computed structures of core eukaryotic protein complexes.
\newblock \emph{Science}, 374\penalty0 (6573):\penalty0 eabm4805, 2021.

\bibitem[Ingraham et~al.(2019)Ingraham, Garg, Barzilay, and Jaakkola]{ingraham2019generative}
Ingraham, J., Garg, V., Barzilay, R., and Jaakkola, T.
\newblock Generative models for graph-based protein design.
\newblock \emph{Advances in neural information processing systems}, 32, 2019.

\bibitem[Jin et~al.(2022)Jin, Wohlwend, Barzilay, and Jaakkola]{jiniterative}
Jin, W., Wohlwend, J., Barzilay, R., and Jaakkola, T.~S.
\newblock Iterative refinement graph neural network for antibody sequence-structure co-design.
\newblock In \emph{International Conference on Learning Representations}, 2022.

\bibitem[Kabsch \& Sander(1983)Kabsch and Sander]{kabsch1983dictionary}
Kabsch, W. and Sander, C.
\newblock Dictionary of protein secondary structure: pattern recognition of hydrogen-bonded and geometrical features.
\newblock \emph{Biopolymers: Original Research on Biomolecules}, 22\penalty0 (12):\penalty0 2577--2637, 1983.

\bibitem[Kingma(2014)]{kingma2014adam}
Kingma, D.~P.
\newblock Adam: A method for stochastic optimization.
\newblock \emph{arXiv preprint arXiv:1412.6980}, 2014.

\bibitem[Li et~al.(2022)Li, Thickstun, Gulrajani, Liang, and Hashimoto]{li2022diffusion}
Li, X., Thickstun, J., Gulrajani, I., Liang, P.~S., and Hashimoto, T.~B.
\newblock Diffusion-lm improves controllable text generation.
\newblock \emph{Advances in Neural Information Processing Systems}, 35:\penalty0 4328--4343, 2022.

\bibitem[Lin et~al.(2022)Lin, Akin, Rao, Hie, Zhu, Lu, Smetanin, dos Santos~Costa, Fazel-Zarandi, Sercu, Candido, et~al.]{lin2022language}
Lin, Z., Akin, H., Rao, R., Hie, B., Zhu, Z., Lu, W., Smetanin, N., dos Santos~Costa, A., Fazel-Zarandi, M., Sercu, T., Candido, S., et~al.
\newblock Language models of protein sequences at the scale of evolution enable accurate structure prediction.
\newblock \emph{bioRxiv}, 2022.

\bibitem[Lin et~al.(2023)Lin, Akin, Rao, Hie, Zhu, Lu, Smetanin, Verkuil, Kabeli, Shmueli, et~al.]{lin2023evolutionary}
Lin, Z., Akin, H., Rao, R., Hie, B., Zhu, Z., Lu, W., Smetanin, N., Verkuil, R., Kabeli, O., Shmueli, Y., et~al.
\newblock Evolutionary-scale prediction of atomic-level protein structure with a language model.
\newblock \emph{Science}, 379\penalty0 (6637):\penalty0 1123--1130, 2023.

\bibitem[Nelson et~al.(2010)Nelson, Dhimolea, and Reichert]{nelson2010development}
Nelson, A.~L., Dhimolea, E., and Reichert, J.~M.
\newblock Development trends for human monoclonal antibody therapeutics.
\newblock \emph{Nature reviews drug discovery}, 9\penalty0 (10):\penalty0 767--774, 2010.

\bibitem[Nichol \& Dhariwal(2021)Nichol and Dhariwal]{nichol2021improved}
Nichol, A.~Q. and Dhariwal, P.
\newblock Improved denoising diffusion probabilistic models.
\newblock In \emph{International conference on machine learning}, pp.\  8162--8171. PMLR, 2021.

\bibitem[Pacesa et~al.(2024)Pacesa, Nickel, Schellhaas, Schmidt, Pyatova, Kissling, Barendse, Choudhury, Kapoor, Alcaraz-Serna, et~al.]{pacesa2024bindcraft}
Pacesa, M., Nickel, L., Schellhaas, C., Schmidt, J., Pyatova, E., Kissling, L., Barendse, P., Choudhury, J., Kapoor, S., Alcaraz-Serna, A., et~al.
\newblock Bindcraft: one-shot design of functional protein binders.
\newblock \emph{bioRxiv}, pp.\  2024--09, 2024.

\bibitem[Satorras et~al.(2021)Satorras, Hoogeboom, and Welling]{satorras2021n}
Satorras, V.~G., Hoogeboom, E., and Welling, M.
\newblock E (n) equivariant graph neural networks.
\newblock In \emph{International conference on machine learning}, pp.\  9323--9332. PMLR, 2021.

\bibitem[Silva et~al.(2019)Silva, Yu, Ulge, Spangler, Jude, Lab{\~a}o-Almeida, Ali, Quijano-Rubio, Ruterbusch, Leung, et~al.]{silva2019novo}
Silva, D.-A., Yu, S., Ulge, U.~Y., Spangler, J.~B., Jude, K.~M., Lab{\~a}o-Almeida, C., Ali, L.~R., Quijano-Rubio, A., Ruterbusch, M., Leung, I., et~al.
\newblock De novo design of potent and selective mimics of il-2 and il-15.
\newblock \emph{Nature}, 565\penalty0 (7738):\penalty0 186--191, 2019.

\bibitem[Sohl-Dickstein et~al.(2015)Sohl-Dickstein, Weiss, Maheswaranathan, and Ganguli]{sohl2015deep}
Sohl-Dickstein, J., Weiss, E., Maheswaranathan, N., and Ganguli, S.
\newblock Deep unsupervised learning using nonequilibrium thermodynamics.
\newblock In \emph{International conference on machine learning}, pp.\  2256--2265. PMLR, 2015.

\bibitem[Song \& Ermon(2019)Song and Ermon]{song2019generative}
Song, Y. and Ermon, S.
\newblock Generative modeling by estimating gradients of the data distribution.
\newblock \emph{Advances in neural information processing systems}, 32, 2019.

\bibitem[Song et~al.(2024{\natexlab{a}})Song, Huang, Li, and Jin]{songsurfpro}
Song, Z., Huang, T., Li, L., and Jin, W.
\newblock Surfpro: Functional protein design based on continuous surface.
\newblock In \emph{Forty-first International Conference on Machine Learning}, 2024{\natexlab{a}}.

\bibitem[Song et~al.(2024{\natexlab{b}})Song, Zhao, Shi, Jin, Yang, and Li]{songgenerative}
Song, Z., Zhao, Y., Shi, W., Jin, W., Yang, Y., and Li, L.
\newblock Generative enzyme design guided by functionally important sites and small-molecule substrates.
\newblock In \emph{Forty-first International Conference on Machine Learning}, 2024{\natexlab{b}}.

\bibitem[Steinegger \& S{\"o}ding(2017)Steinegger and S{\"o}ding]{steinegger2017mmseqs2}
Steinegger, M. and S{\"o}ding, J.
\newblock Mmseqs2 enables sensitive protein sequence searching for the analysis of massive data sets.
\newblock \emph{Nature biotechnology}, 35\penalty0 (11):\penalty0 1026--1028, 2017.

\bibitem[Stern et~al.(2013)Stern, Case, and Hackel]{stern2013alternative}
Stern, L.~A., Case, B.~A., and Hackel, B.~J.
\newblock Alternative non-antibody protein scaffolds for molecular imaging of cancer.
\newblock \emph{Current opinion in chemical engineering}, 2\penalty0 (4):\penalty0 425--432, 2013.

\bibitem[Varadi et~al.(2022)Varadi, Anyango, Deshpande, Nair, Natassia, Yordanova, Yuan, Stroe, Wood, Laydon, et~al.]{varadi2022alphafold}
Varadi, M., Anyango, S., Deshpande, M., Nair, S., Natassia, C., Yordanova, G., Yuan, D., Stroe, O., Wood, G., Laydon, A., et~al.
\newblock Alphafold protein structure database: massively expanding the structural coverage of protein-sequence space with high-accuracy models.
\newblock \emph{Nucleic acids research}, 50\penalty0 (D1):\penalty0 D439--D444, 2022.

\bibitem[Vaswani(2017)]{vaswani2017attention}
Vaswani, A.
\newblock Attention is all you need.
\newblock \emph{Advances in Neural Information Processing Systems}, 2017.

\bibitem[V{\'a}zquez~Torres et~al.(2024)V{\'a}zquez~Torres, Leung, Venkatesh, Lutz, Hink, Huynh, Becker, Yeh, Juergens, Bennett, et~al.]{vazquez2024novo}
V{\'a}zquez~Torres, S., Leung, P.~J., Venkatesh, P., Lutz, I.~D., Hink, F., Huynh, H.-H., Becker, J., Yeh, A. H.-W., Juergens, D., Bennett, N.~R., et~al.
\newblock De novo design of high-affinity binders of bioactive helical peptides.
\newblock \emph{Nature}, 626\penalty0 (7998):\penalty0 435--442, 2024.

\bibitem[Wang et~al.(2024)Wang, Zheng, Fei, Xue, Huang, and Gu]{wangdiffusion}
Wang, X., Zheng, Z., Fei, Y., Xue, D., Huang, S., and Gu, Q.
\newblock Diffusion language models are versatile protein learners.
\newblock In \emph{Forty-first International Conference on Machine Learning}, 2024.

\bibitem[Watson et~al.(2023)Watson, Juergens, Bennett, Trippe, Yim, Eisenach, Ahern, Borst, Ragotte, Milles, et~al.]{watson2023novo}
Watson, J.~L., Juergens, D., Bennett, N.~R., Trippe, B.~L., Yim, J., Eisenach, H.~E., Ahern, W., Borst, A.~J., Ragotte, R.~J., Milles, L.~F., et~al.
\newblock De novo design of protein structure and function with rfdiffusion.
\newblock \emph{Nature}, 620\penalty0 (7976):\penalty0 1089--1100, 2023.

\bibitem[Wu et~al.(2024)Wu, Jiang, Hicks, Liu, Muratspahi{\'c}, Ramelot, Liu, McNally, Gaur, Coventry, et~al.]{wu2024sequence}
Wu, K., Jiang, H., Hicks, D.~R., Liu, C., Muratspahi{\'c}, E., Ramelot, T.~A., Liu, Y., McNally, K., Gaur, A., Coventry, B., et~al.
\newblock Sequence-specific targeting of intrinsically disordered protein regions.
\newblock \emph{bioRxiv}, pp.\  2024--07, 2024.

\bibitem[Wu et~al.(2022)Wu, Gong, Liu, Ye, and Liu]{wu2022diffusion}
Wu, L., Gong, C., Liu, X., Ye, M., and Liu, Q.
\newblock Diffusion-based molecule generation with informative prior bridges.
\newblock \emph{Advances in Neural Information Processing Systems}, 35:\penalty0 36533--36545, 2022.

\bibitem[Yan et~al.(2020)Yan, Tao, He, and Huang]{yan2020hdock}
Yan, Y., Tao, H., He, J., and Huang, S.-Y.
\newblock The hdock server for integrated protein--protein docking.
\newblock \emph{Nature protocols}, 15\penalty0 (5):\penalty0 1829--1852, 2020.

\bibitem[Yang et~al.(2024)Yang, Hicks, Ghosh, Schwartze, Conventry, Goreshnik, Allen, Halabiya, Kim, Hinck, et~al.]{yang2024design}
Yang, W., Hicks, D.~R., Ghosh, A., Schwartze, T.~A., Conventry, B., Goreshnik, I., Allen, A., Halabiya, S.~F., Kim, C.~J., Hinck, C.~S., et~al.
\newblock Design of high affinity binders to convex protein target sites.
\newblock \emph{bioRxiv}, 2024.

\bibitem[Yim et~al.(2023)Yim, Trippe, De~Bortoli, Mathieu, Doucet, Barzilay, and Jaakkola]{yim2023se}
Yim, J., Trippe, B.~L., De~Bortoli, V., Mathieu, E., Doucet, A., Barzilay, R., and Jaakkola, T.
\newblock Se (3) diffusion model with application to protein backbone generation.
\newblock In \emph{International Conference on Machine Learning}, pp.\  40001--40039. PMLR, 2023.

\bibitem[Zambaldi et~al.(2024)Zambaldi, La, Chu, Patani, Danson, Kwan, Frerix, Schneider, Saxton, Thillaisundaram, et~al.]{zambaldi2024novo}
Zambaldi, V., La, D., Chu, A.~E., Patani, H., Danson, A.~E., Kwan, T.~O., Frerix, T., Schneider, R.~G., Saxton, D., Thillaisundaram, A., et~al.
\newblock De novo design of high-affinity protein binders with alphaproteo.
\newblock \emph{arXiv preprint arXiv:2409.08022}, 2024.

\bibitem[Zhang \& Skolnick(2004)Zhang and Skolnick]{zhang2004scoring}
Zhang, Y. and Skolnick, J.
\newblock Scoring function for automated assessment of protein structure template quality.
\newblock \emph{Proteins: Structure, Function, and Bioinformatics}, 57\penalty0 (4):\penalty0 702--710, 2004.

\end{thebibliography}
\bibliographystyle{icml2025}

\newpage
\appendix
\onecolumn
\section*{Appendix}

\section{Data Statistics}
\subsection{Data Statistics for \dataset}
\label{Appendix: data_statistics_procomplexbench}
Detailed data statistics for the curated protein-protein complex dataset are reported in Table~\ref{Tab: data_statistics}.

\begin{table}[ht]
\centering
\begin{tabular}{lcc}
\midrule
 & Cluster & Complex  \\
\midrule
Training &22,827 &706,149  \\
Validation  & 10&155\\
Test & 10& 56\\
Total & 22,847& 706,360\\
\bottomrule
\end{tabular}
\vspace{-0.5em}
\caption{Data statistics for \dataset.}
\label{Tab: data_statistics}
\end{table}

\begin{table}[ht]
\begin{center}
\begin{tabular}{lcccccccccc}
\midrule
\multirow{2}{*}{Models} & \multicolumn{5}{c}{Seen Class} & \multicolumn{5}{c}{Zero-Shot} \\
\cmidrule(r){2-6} \cmidrule(r){7-11} 
& FGFR2	&InsulinR&	PDGFR&	TGFb&	VirB8&	H3&	IL7Ra&	EGFR&	TrkA&	Tie2  \\
\midrule
Training &465&192&210	&77	&57&--&	--&--&	--&	-- \\
Validation&57&23&26	&9&7&--&--&	--&--&--\\
Test&57	&23&26&	9&7&38&	7&4&4&2\\
Total& 579&238&262&95&71&38&7&4&4&	2\\
\bottomrule
\end{tabular}
\end{center}
\caption{Detailed data statistics for curated target-protein mini-binder complex dataset.}
\label{Table:target-protein mini-binder dataset statistics}
\end{table}

\subsection{Data Statistics for Target-Protein Mini-Binder Complex Dataset}
\label{Appendix: data_statistics_for_target_protein_mini_binder}
We collect experimentally confirmed positive target-protein mini-binder complexes against ten target proteins with diverse structural properties from \citet{bennett2023improving}. The detailed data statistics for the ten target proteins are reported in Table~\ref{Table:target-protein mini-binder dataset statistics}.

\begin{table}[ht]
\centering
\begin{tabular}{lcccc}
\midrule
 & CDR-H1 Clustering & CDR-H2 Clustering & CDR-H3 Clustering \\
\midrule
Complexes &4,261	&4,261&	4261 \\
Clusters  & 1,034&	1,418&	2,254\\
Training & 3,590&	3,576&	3,427\\
Validation &403&	399&	451 \\
Test & 268&	286&	383\\
\bottomrule
\end{tabular}
\vspace{-0.5em}
\caption{Detailed data statistics for antigen-antibody complex dataset.}
\label{Tab: data_statistics_antibody_antigen_complex}
\end{table}

\subsection{Data Statistics for Antigen-Antibody Complex Dataset}
\label{Appendix:antibody-antigen-complex-data-statistics}
The detailed data statistics for each CDR clustering are reported in Table~\ref{Tab: data_statistics_antibody_antigen_complex}.

\begin{table*}[!t]
\footnotesize
\begin{center}
\setlength{\tabcolsep}{0.88mm}{
\begin{tabular}{lcccccccc}
\midrule
Methods  & ipTM ($\uparrow$) & pTM ($\uparrow$) & PAE ($\downarrow$) & pLDDT($\uparrow$) & Success Rate ($\uparrow$) & H1 Novelty ($\uparrow$) & H2 Novelty ($\uparrow$) & H3 Novelty ($\uparrow$)\\
\midrule 
SEnc +ProteinMPNN&0.364&	0.566&	15.021&	76.063&	4.85\%&	60.37\%&	60.95\%&	69.39\%\\
InterleavingDiff &0.426&	0.621&	14.993&	81.581&	7.83\%&	\textbf{62.38\%}&	68.21\%&	77.48\%\\
\network Network&\textbf{0.501}&	0.636&	\textbf{12.759}&	\textbf{82.608}&	\textbf{14.55\%}&	34.87\%&	38.50\%&	57.55\%\\
\rowcolor{myblue}
\model & 0.484	&\textbf{0.638}	&14.058	&82.509	&11.11\%&	61.62\%&	\textbf{69.69\%}&	\textbf{77.62\%}\\
\bottomrule
\end{tabular}}
\end{center}
\caption{Model performance on antigen-antibody complex design task for CDR-H1 clustering. ($\uparrow$): the higher the better. ($\downarrow$): the lower the better. Novelty scores for CDR-H1, CDR-H2, and CDR-H3 are reported as H1, H2, and H3 novelty, respectively. \model achieves the highest pTM and novelty scores across the designed CDR-H2 and CDR-H3.}
\label{Table:antibody_antigen_complex_design_cdrh1}
\end{table*}

\begin{table*}[!t]
\footnotesize
\begin{center}
\setlength{\tabcolsep}{0.88mm}{
\begin{tabular}{lcccccccc}
\midrule
Methods  & ipTM ($\uparrow$) & pTM ($\uparrow$) & PAE ($\downarrow$) & pLDDT($\uparrow$) & Success Rate ($\uparrow$) & H1 Novelty ($\uparrow$) & H2 Novelty ($\uparrow$) & H3 Novelty ($\uparrow$)\\
\midrule
SEnc +ProteinMPNN&0.525	&0.629&	13.339	&79.260&	7.34\%&	\textbf{62.23\%}&	59.71\%&	74.66\%\\
InterleavingDiff &0.539&	0.681&	12.546&	81.155&	17.48\%&	55.54\%&	59.03\%&	74.51\%\\
\network Network&0.547	&0.658&	11.929&	82.829&	20.27\%	&30.29\%&	39.94\%&	61.13\%\\
\rowcolor{myblue}
\model & \textbf{0.561}&	\textbf{0.682}	&\textbf{12.535}	&\textbf{83.162}	&\textbf{22.03\%}&	57.77\%&	\textbf{65.31\%}&	\textbf{75.95\%}\\
\bottomrule
\end{tabular}}
\end{center}
\caption{Model performance on antigen-antibody complex design task for CDR-H2 clustering. ($\uparrow$): the higher the better. ($\downarrow$): the lower the better. Novelty scores for CDR-H1, CDR-H2, and CDR-H3 are reported as H1, H2, and H3 novelty, respectively. \model achieves the highest scores on almost all metrics except CDR-H1 novelty score.}
\label{Table:antibody_antigen_complex_design_cdrh2}
\end{table*}

\begin{table*}[!t]
\footnotesize
\begin{center}
\setlength{\tabcolsep}{0.88mm}{
\begin{tabular}{lcccccccc}
\midrule
Methods  & ipTM ($\uparrow$) & pTM ($\uparrow$) & PAE ($\downarrow$) & pLDDT($\uparrow$) & Success Rate ($\uparrow$) & H1 Novelty ($\uparrow$) & H2 Novelty ($\uparrow$) & H3 Novelty ($\uparrow$)\\
\midrule
SEnc +ProteinMPNN&0.457	&0.539&	14.894&	78.036&	6.52\%&	\textbf{62.75\%}&	57.93\%&	68.49\%\\
InterleavingDiff &0.523&	 0.659	&13.210&	80.082&	10.70\%&	54.69\%&	63.53\%&	74.75\%\\
\network Network&0.569&	0.668&	\textbf{11.506}&	82.682	&15.40\%&	26.16\%&	34.67\%&	60.91\%\\
\rowcolor{myblue}
\model &\textbf{0.574}&	\textbf{0.683}&	12.405	&\textbf{82.810}&	\textbf{17.52\%}&	53.98\%&	\textbf{64.18\%}&	\textbf{74.94\%}\\
\bottomrule
\end{tabular}}
\end{center}
\caption{Model performance on antigen-antibody complex design task for CDR-H3 clustering. ($\uparrow$): the higher the better. ($\downarrow$): the lower the better. Novelty scores for CDR-H1, CDR-H2, and CDR-H3 are reported as H1, H2, and H3 novelty, respectively. \model achieves the highest scores on the antigen-antibody binding affinity~(ipTM) and success rate.}
\label{Table:antibody_antigen_complex_design_cdrh3}
\end{table*}

\section{Supplementary Experimental Results}
\subsection{Antigen-Antibody Complex Design Performance for Each CDR Clustering}
\label{Appendix:addtional-result-antibody-antigen}
Model performance for each CDR clustering is provided in Table~\ref{Table:antibody_antigen_complex_design_cdrh1}, ~\ref{Table:antibody_antigen_complex_design_cdrh2} and ~\ref{Table:antibody_antigen_complex_design_cdrh3}. Our proposed \model achieves higher success rate than the strongest baseline \network Network on average and obtains much higher novelty scores.

\begin{table*}[!t]
\footnotesize
\begin{center}
\setlength{\tabcolsep}{1.8mm}{
\begin{tabular}{llccccccc}
\midrule
\multicolumn{2}{c}{Methods}  & ipTM ($\uparrow$) & pTM ($\uparrow$) & PAE ($\downarrow$) & pLDDT($\uparrow$) & Success Rate ($\uparrow$) & Novelty ($\uparrow$) & Diversity ($\uparrow$) \\
\midrule
\multicolumn{2}{c}{Ground Truth} &0.691& 0.782 & 7.901& 86.987&69.64\%&-- &-- \\
\hdashline
\multirow{2}{*}{top-1}&\cellcolor{myblue}\model &\cellcolor{myblue}\textbf{0.700}&\cellcolor{myblue}\textbf{0.779}&\cellcolor{myblue}\textbf{9.153}&\cellcolor{myblue}\textbf{83.765}&\cellcolor{myblue}50.00\%&\cellcolor{myblue}\textbf{89.46\%}&\cellcolor{myblue}-- \\
& ~~- With pretraining&0.666&	0.770&	9.647&	82.952&\textbf{53.57\%}	&88.89\%& --\\
\midrule
\multirow{2}{*}{top-5}&\cellcolor{myblue}\model &\cellcolor{myblue}\textbf{0.665}&\cellcolor{myblue}\textbf{0.747}&\cellcolor{myblue}10.231&\cellcolor{myblue}\textbf{81.659}&\cellcolor{myblue}\textbf{45.71\%}&\cellcolor{myblue}88.93\%&\cellcolor{myblue}90.76\% \\
& ~~- With pretraining&0.663&0.733&	\textbf{10.087}&	79.597&	35.36\%&	\textbf{89.90\%}&\textbf{	91.81\%}\\
\midrule
\multirow{2}{*}{top-10}&\cellcolor{myblue}\model &\cellcolor{myblue}\textbf{0.633}&\cellcolor{myblue}\textbf{0.729}&\cellcolor{myblue}\textbf{10.895}&\cellcolor{myblue}\textbf{80.322}&\cellcolor{myblue}\textbf{37.68\%}&\cellcolor{myblue}89.10\%&\cellcolor{myblue}91.09\% \\
& ~~- With pretraining&0.626&0.710&	11.255&78.068&	27.68\%&	\textbf{90.74\%}&	\textbf{92.17\%}\\
\bottomrule
\end{tabular}}
\end{center}
\caption{Model performance on additional Swiss-Prot data pretraining. ($\uparrow$): the higher the better. ($\downarrow$): the lower the better. Calculating the diversity score for the top-1 candidate is unnecessary, as it consists of only a single candidate.}
\label{Table:siwss-prot-pretraining}
\end{table*}

\begin{table*}[!t]
\footnotesize
\begin{center}
\setlength{\tabcolsep}{1.8mm}{
\begin{tabular}{llccccccc}
\midrule
\multicolumn{2}{c}{Methods}  & ipTM ($\uparrow$) & pTM ($\uparrow$) & PAE ($\downarrow$) & pLDDT($\uparrow$) & Success Rate ($\uparrow$) & Novelty ($\uparrow$) & Diversity ($\uparrow$) \\
\midrule
\multicolumn{2}{c}{Ground Truth} &0.691& 0.782 & 7.901& 86.987&69.64\%&-- &-- \\
\hdashline
\multirow{4}{*}{top-1}&SEnc +ProteinMPNN&0.586&	0.693&	12.029	&77.064	&28.57\%&	63.28\%&-- \\
& InterleavingDiff &0.631&	0.722&	11.199	&79.651	&33.92\%&	\textbf{91.61\%} & --\\
&\network Network&0.634&	0.736&	\textbf{10.256}	&81.461&	37.50\%&	37.60\%& --\\
&\cellcolor{myblue}\model &\cellcolor{myblue}\textbf{0.644}&\cellcolor{myblue}\textbf{0.739}&\cellcolor{myblue}10.289&\cellcolor{myblue}\textbf{81.926}&\cellcolor{myblue}\textbf{39.28\%}&\cellcolor{myblue}88.76\%&\cellcolor{myblue}-- \\
\midrule
\multirow{4}{*}{top-5}&SEnc +ProteinMPNN&0.569	&0.683&	12.931	&76.201	&14.28\%&	65.37\%&	62.21\%\\
& InterleavingDiff &0.594&	0.695	&12.594	&76.465	&15.00\%&	\textbf{91.69\%}	&\textbf{92.09\%}\\
&\network Network&0.602&	0.709&	11.528&	\textbf{80.800}	&\textbf{32.14\%}&	37.26\%&	15.04\% \\
&\cellcolor{myblue}\model &\cellcolor{myblue}\textbf{0.606}&\cellcolor{myblue}\textbf{0.713}&\cellcolor{myblue}\textbf{11.459}&\cellcolor{myblue}79.176&\cellcolor{myblue}30.36\%&\cellcolor{myblue}89.11\%&\cellcolor{myblue}91.45\% \\
\midrule
\multirow{4}{*}{top-10}&SEnc +ProteinMPNN&0.553&	0.662&	13.641	&75.148&	7.14\%&	67.89\%&	63.92\% \\
& InterleavingDiff &0.581	&0.677	&13.512&	74.090&	7.50\%&	\textbf{91.60\%}&	\textbf{92.49\%}\\
&\network Network&0.586&	0.691&	\textbf{11.856}&	\textbf{78.920}&	\textbf{21.42\%}&	37.43\%&	15.18\% \\
&\cellcolor{myblue}\model &\cellcolor{myblue}\textbf{0.592}&\cellcolor{myblue}\textbf{0.695}&\cellcolor{myblue}12.394&\cellcolor{myblue}77.014&\cellcolor{myblue}18.21\%&\cellcolor{myblue}90.18\%&\cellcolor{myblue}91.68\% \\
\bottomrule
\end{tabular}}
\end{center}
\caption{Model performance with candidate pool size set to 30. ($\uparrow$): the higher the better. ($\downarrow$): the lower the better. Calculating the diversity score for the top-1 candidate is unnecessary, as it consists of only a single candidate. }
\label{Table:candidate-size-30}
\end{table*}

\subsection{Swiss-Prot Data Pretraining}
\label{Appendix: Siwss-Prot-pretraining}
To evaluate if pretraining our \model on additional data will improve model performance, we first pretrain our model on curated pseudo Swiss-Prot complexes, and then continue training the model on our constructed \dataset. We compare this model and our original \model in Table~\ref{Table:siwss-prot-pretraining}. It shows pretraining on Swiss-Prot data does not provide significant benefit.

\subsection{Smaller Candidate Pool Size}
\label{Appendix: candidate-pool-size}
To study the effect of total candidate size, we provide the results for a candidate pool size of 30 in Table~\ref{Table:candidate-size-30}. It shows reducing the candidate pool size leads to worse model performance on almost all metrics.

\section{Additional Experimental Details}
\subsection{\dataset Construction Details}
\label{Appendix:protein_protein_complex curation_detail}
We curate a large-scale protein-protein complex dataset by identifying chain-pair interfaces, following the data processing pipeline introduced in AlphaFold3~\cite{abramson2024accurate}. The curation process begins by applying a series of quality filters to identify valid PDB entries:
\begin{itemize}
    \item Retain structures with a reported resolution of 9Å or better.
    \item Remove polymer chains containing unknown residues.
    \item Exclude protein chains where consecutive $C_\alpha$ atoms are separated by more than 10Å.
    \item Filter out polymer chains with fewer than four resolved residues.
    \item Eliminate clashing chains, defined as chains with more than 30\% of their atoms located within 1.7Å of an atom in another chain. When two chains clashed:
    \begin{itemize}
        \item The chain with the higher percentage of clashing atoms is removed.
        \item If the clashing percentage is equal, the chain with fewer total atoms is removed.
        \item If both chains have the same number of atoms, the chain with the larger chain ID is removed.
    \end{itemize}
\end{itemize}
This filtering process result in 126,569 valid PDB entries. From these entries, chain-pair interfaces are identified, defined as pairs of chains with a minimum heavy atom separation of less than 5Å. This step yields a total of 367,016 chain-pair interfaces.

\subsection{Model Training Details}
\label{Appendix: model_training_details}
The embedding dimensionality is configured at 1,280. \model is trained for 1,000,000 steps on a single NVIDIA RTX A6000 GPU using the Adam optimizer~\cite{kingma2014adam}. The batch size is set to 1,024 tokens, and the learning rate is initialized at 5e-6
 . For structure diffusion, the starting and ending values of $\beta$ are set to  1e-7 and 2e-3, respectively, with a variance schedule of 2. The cosine schedule offset for sequence diffusion is configured at 0.01. The number of $k$-nearest neighbors is fixed at 32. Additionally, a learning rate warm-up is applied over the first 4,000 steps to stabilize the training process.

 \subsection{Detailed Explanation of Metrics}
\label{Appendix:metrics_explanation}
We provide the detailed explanation of metrics as follows:
\begin{itemize}
    \item \textbf{ipTM} is an interfacial variant of the predicted TM-score, evaluating the interaction between different chains~\cite{abramson2024accurate}. Values higher than 0.8 represent confident high-quality predictions.
    \item \textbf{pTM} is a confidence score estimating the accuracy of global protein structure prediction. A pTM score above 0.5 means the overall predicted fold for the complex might be similar to the true structure~\cite{zhang2004scoring}.
    \item \textbf{PAE} estimates the error in the relative position and orientation between two tokens in the predicted structure. Higher values indicate higher predicted error and therefore lower confidence. 
    \item \textbf{pLDDT} aims to predict a modified LDDT score that only considers distances to polymers. It is a per-atom confidence estimate on a 0-100 scale where a higher value indicates higher structural stability.
\end{itemize}

\subsection{Target-Protein Mini-Binder Complex Design Evaluation Details}
\label{Appendix: evaluation-detail-of-target-protein-minibinder-complex}
To calculate the pAE\_interaction score, we first predict the structure of the designed binder sequence using ESMFold~\cite{lin2023evolutionary}. The predicted structure is then superimposed onto the experimentally validated positive binder, and the AF2 pAE\_interaction score is calculated for the resulting target-protein mini-binder complex. Using these scores, we determine the \textbf{success rate} by generating five mini-binder candidates for each target protein. The success rate is defined as the proportion of designed binders that achieve a lower pAE\_interaction score with the target protein than the ground truth positive binder:
\begin{equation}
\small 
\mathrm{SR}=\frac{1}{|\mathcal{D}_{\text{test}}| * k}\sum_{\mathcal{T} \in \mathcal{D}_{\text{test}}} \sum_{i=1}^k \mathrm{II}(f(\mathcal{T}, \mathcal{B}_i, \mathcal{B}))
\end{equation}
where $k$=5, $f(\mathcal{T}, \mathcal{B}_i, \mathcal{B})=\mathrm{true}$ if the $i$-th binder candidate $\mathcal{B}_i$ for target protein $\mathcal{T}$ achieves a lower pAE\_interaction score than the positive binder $\mathcal{B}$. Otherwise, $f(\mathcal{T}, \mathcal{B}_i, \mathcal{B})=\mathrm{false}$.

\section{Additional Designed Cases}
\label{Appendix: case_study}
\begin{figure*}
\begin{minipage}[t]{0.25\linewidth}
\centering
\includegraphics[width=4.0cm]{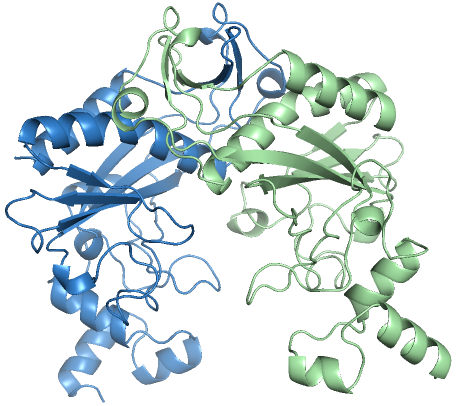}
{\small{(a) ipTM=0.86, pTM=0.86, \\PAE=6.93, pLDDT=87.83, novelty=83.19\%}}
\end{minipage}%
\begin{minipage}[t]{0.25\linewidth}
\centering
\includegraphics[width=4.05cm]{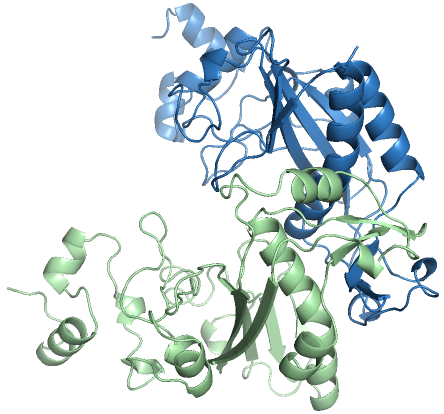}
\small{(b) ipTM=0.85, pTM=0.85, \\PAE=7.07, pLDDT=87.61, novelty=96.72\%}
\end{minipage}%
\begin{minipage}[t]{0.25\linewidth}
\centering
\includegraphics[width=3.55cm]{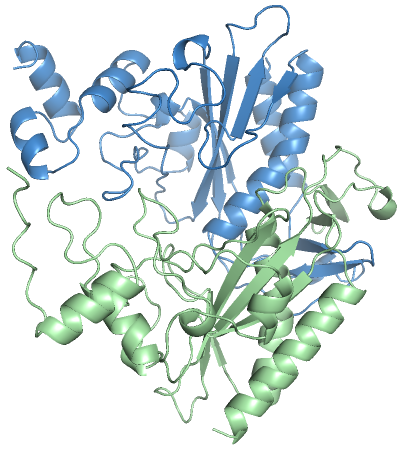}
\small{(c) ipTM=0.82, pTM=0.83, PAE=7.49, pLDDT=88.31, novelty=81.55\%}
\end{minipage}%
\begin{minipage}[t]{0.25\linewidth}
\centering
\includegraphics[width=3.55cm]{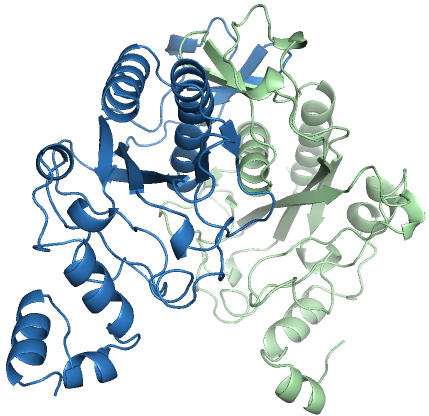}
\small{(d) ipTM=0.85, pTM=0.86, PAE=6.93, pLDDT=88.50, novelty=93.44\%}
\end{minipage}
\begin{minipage}[t]{0.25\linewidth}
\centering
\includegraphics[width=4.05cm]{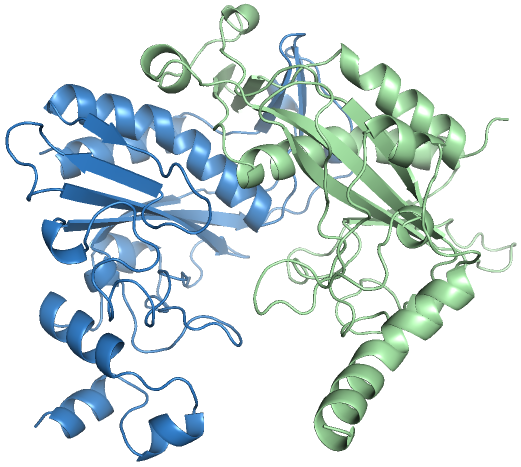}
\small{(e) ipTM=0.89, pTM=0.88, PAE=6.53, pLDDT=88.67, novelty=96.66\%}
\end{minipage}%
\begin{minipage}[t]{0.25\linewidth}
\centering
\includegraphics[width=5.05cm]{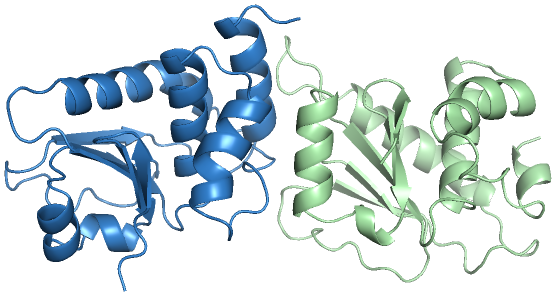}
\small{(f) ipTM=0.76, pTM=0.81, PAE=8.25, pLDDT=82.33, novelty=95.36\%}
\end{minipage}%
\begin{minipage}[t]{0.25\linewidth}
\centering
\includegraphics[width=3.8cm]{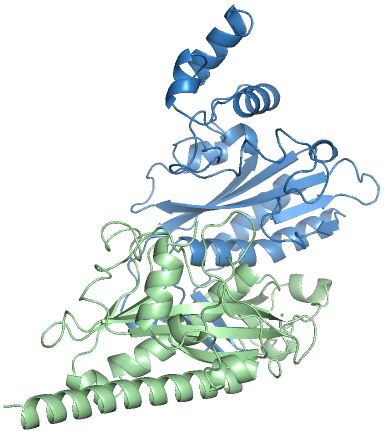}
\small{(g) ipTM=0.87, pTM=0.86, PAE=6.86, pLDDT=88.10, novelty=94.37\%}
\end{minipage}%
\begin{minipage}[t]{0.25\linewidth}
\centering
\includegraphics[width=3.8cm]{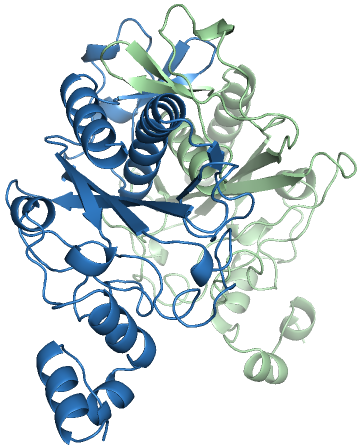}
\small{(c) ipTM=0.86, pTM=0.85, PAE=6.40, pLDDT=89.75, novelty=92.21\%}
\end{minipage}
\vspace{-0.8em}
	\caption{Designed complexes for general protein-protein complex design by our \model. All of them achieve an ipTM score approaching or higher than 0.8, pTM score above 0.8, PAE lower than 10 and pLDDT better than 80. These designed binder sequences also have novelty scores higher than 80\%, validating that \model is capable of designing novel and high-affinity protein-binding proteins across diverse protein targets.} 
 \label{Fig: general-protein-complex-design-case-study}
\end{figure*}

\begin{figure*}
\begin{minipage}[t]{0.25\linewidth}
\centering
\includegraphics[width=4.0cm]{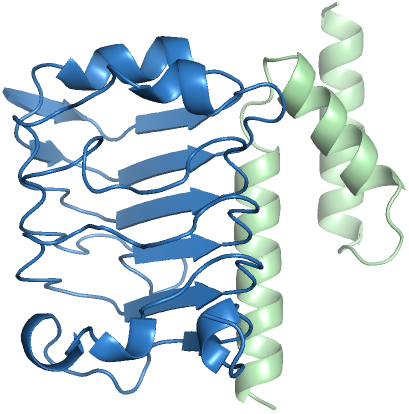}
{\small{(a) ipTM=0.8, pTM=0.87, \\PAE=5.24, pLDDT=85.87, novelty=91.52\%, target protein=H3}}
\end{minipage}%
\begin{minipage}[t]{0.25\linewidth}
\centering
\includegraphics[width=4.05cm]{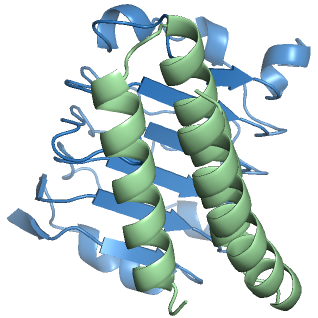}
\small{(b) ipTM=0.78, pTM=0.88, \\PAE=4.37, pLDDT=87.47, novelty=93.84\%, target protein=IL7Ra}
\end{minipage}%
\begin{minipage}[t]{0.25\linewidth}
\centering
\includegraphics[width=3.55cm]{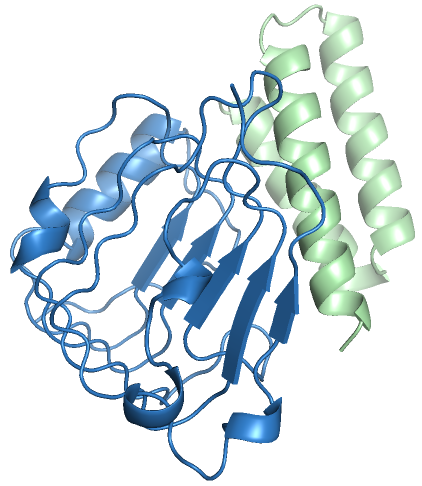}
\small{(c) ipTM=0.72, pTM=0.79, PAE=8.40, pLDDT=80.70, novelty=91.93\%, target protein=IL7Ra}
\end{minipage}%
\begin{minipage}[t]{0.25\linewidth}
\centering
\includegraphics[width=3.55cm]{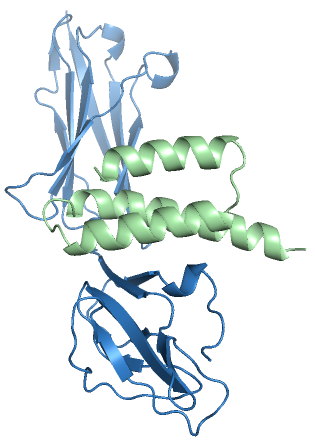}
\small{(d) ipTM=0.77, pTM=0.74, PAE=7.34, pLDDT=84.39, novelty=92.06\%, target protein=InsulinR}
\end{minipage}
\vspace{-0.8em}
	\caption{Designed complexes by our \model for target-protein mini-binder complex design. All of them achieve an ipTM score higher than 0.7, pTM score above 0.7, PAE lower than 10 and pLDDT better than 80. These designed binder sequences also have novelty scores higher than 80\%, validating that \model is capable of designing novel and high-affinity binder proteins across diverse protein targets.} 
 \label{Fig: target-protein-minibinder-case-study}
\end{figure*}

\begin{figure*}
\begin{minipage}[t]{0.25\linewidth}
\centering
\includegraphics[width=4.0cm]{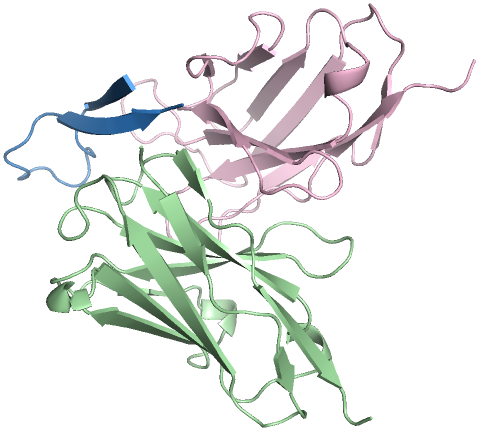}
{\small{(a) ipTM=0.81, pTM=0.87, \\PAE=6.30, pLDDT=84.22, CDR-H1 novelty=62.50\%, CDR-H2 novelty=50.00\%, CDR-H3 novelty=83.33\%}}
\end{minipage}%
\begin{minipage}[t]{0.25\linewidth}
\centering
\includegraphics[width=4.05cm]{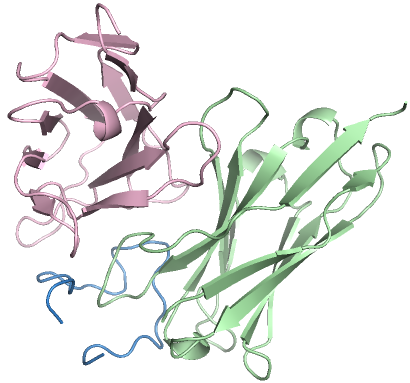}
\small{(b) ipTM=0.83, pTM=0.89, \\PAE=5.36, pLDDT=87.38, CDR-H1 novelty=50.00\%, CDR-H2 novelty=70.00\%, CDR-H3 novelty=83.33\%}
\end{minipage}%
\begin{minipage}[t]{0.25\linewidth}
\centering
\includegraphics[width=4.55cm]{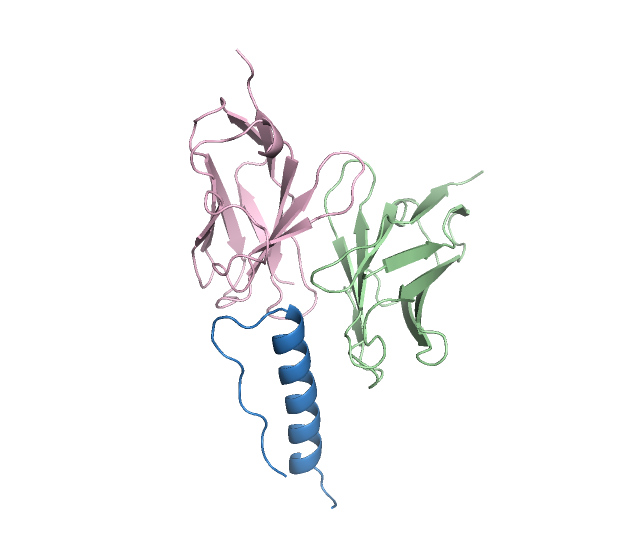}
\small{(c) ipTM=0.89, pTM=0.91, \\PAE=4.01, pLDDT=91.91, CDR-H1 novelty=62.50\%, CDR-H2 novelty=50.00\%, CDR-H3 novelty=60.00\%}
\end{minipage}%
\begin{minipage}[t]{0.25\linewidth}
\centering
\includegraphics[width=3.55cm]{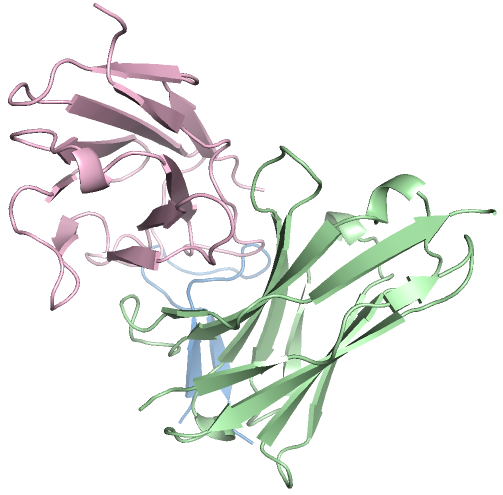}
\small{(d) ipTM=0.81, pTM=0.87, \\PAE=6.09, pLDDT=84.18, CDR-H1 novelty=62.50\%, CDR-H2 novelty=50.00\%, CDR-H3 novelty=75.00\%}
\end{minipage}
\vspace{-0.8em}
	\caption{Designed complexes by our \model for antigen-antibody complex design. All of them achieve an ipTM score higher than 0.8, pTM score above 0.8, PAE lower than 10 and pLDDT better than 80. These designed binder sequences also have CDR-H1, H2 and H3 novelty scores higher than 50\%, validating that \model is capable of designing novel and high-affinity antibodies for antigens.} 
 \label{Fig: antibody-antigen-complex-case-study}
\end{figure*}

We provide more designed protein-protein complexes for general protein-protein complex design, target-protein mini-binder complex design and antigen-antibody complex design in Figure~\ref{Fig: general-protein-complex-design-case-study}, ~\ref{Fig: target-protein-minibinder-case-study} and ~\ref{Fig: antibody-antigen-complex-case-study}, respectively. It shows our \model is capable of designing novel and high-affinity protein-binding proteins for diverse protein targets across a wide range of domains.

\newpage

\end{document}